\def\gmode{} %%use this to include figures

 \def\gdriver{dvipdfmx}%for dvipdfmx to make pdf from dvi
\newcommand{\dtitle}[1]{\title{ \if \gmode \else
\color{red} Demo mode!\\
comment out \textbackslash def \textbackslash gmode\{demo\} at the header to include figures \color{black}\\
\fi
#1 }}
\ifx\pdfoutput\undefined
\else
 \def\gdriver{}
\fi
\documentclass[\gmode,\gdriver,10pt,twocolumn,letterpaper]{article}

\usepackage{iccv}
\usepackage{times}
\usepackage{epsfig}
\usepackage{graphicx}
\usepackage{amsmath}
\usepackage{amssymb}

\usepackage{color}
\newcommand{\fnote}[1]{}
\newcommand{\fnoteIII}[1]{#1}
\newcommand{\knote}[1]{{\color{red} \bf #1 \color{black}}}
\newcommand{\bnote}[1]{{\color{red} \bf #1 \color{black}}}
\newcommand{\mnote}[1]{}

\newcommand{\mcut}[1]{}
\newcommand{\kcut}[1]{}
\newcommand{\fcut}[1]{}
\newcommand{\jptext}[1]{}

\newcommand{\snote}[1]{{\color{red} \bf #1 \color{black}}}

\ifx\pdfoutput\undefined
\else
 \usepackage{CJKutf8}
 \renewcommand{\fnote}[1]{}
 \renewcommand{\knote}[1]{}
 \renewcommand{\mnote}[1]{}
 \renewcommand{\bnote}[1]{}
\fi

\if 0

\newcommand{\etal}{{\it et al. }}
\newcommand{\ie}{{\it i.e.}}
\newcommand{\eg}{{\it e.g.}}

\fi

\iccvfinalcopy % *** Uncomment this line for the final submission

\def\iccvPaperID{2257} % *** Enter the CVPR Paper ID here

\ificcvfinal\fi
\begin{document}

%%%%%%%%% TITLE
\dtitle{
\vspace{-4mm}
Depth estimation using structured light flow\\
%\dtitle{Shape recovery from structured light flow\\
%Structure from motion using projected pattern flow\\ 
--- analysis of projected pattern flow on an object's surface ---
\vspace{-2mm}
}

\author{Ryo Furukawa\\
Hiroshima City University,
%Hiroshima, Japan\\
Japan\\
%{\tt\small ryo-f@hiroshima-cu.ac.jp}
\and
Ryusuke Sagawa\\
%National Institute of Advanced Industrial Science and Technology, \\
AIST, 
%Tsukuba, Japan\\
Japan\\
%{\tt\small ryusuke.sagawa@aist.go.jp}
\and
Hiroshi Kawasaki\\
Kyushu University,
%Fukuoka, Japan\\
Japan\\
%{\tt\small kawasaki@ait.kyushu-u.ac.jp}
}
\graphicspath{{../170721-cvpr-lightflow/}{./}}

\maketitle
%\thispagestyle{empty}

%%%%%%%%% ABSTRACT
\begin{abstract}
\vspace{-2mm}
Shape reconstruction techniques using structured light have been widely    
researched and developed due to their robustness, high precision, 
and density. 
%Among them, a structured light technique is the most successful and 
%    widely used.
Because the techniques are based on decoding a pattern to find correspondences, 
it implicitly requires that the projected patterns be 
    clearly captured by an image sensor, 
    \ie, to avoid defocus and motion blur of the projected pattern. %, and that
%    both the system and the scene should be static.
Although intensive researches have been conducted for solving defocus blur, 
    few researches for motion blur and only solution is to capture 
    with extremely fast shutter speed.
%	the motion of the scene with respect to the system be ignorable during
%	exposure.
%
%If both or either of them move faster then shutter speed, the projected pattern 
%    gets blurred and a captured image is usually considered just failure and thrown away.
In this paper, unlike the previous approaches, we actively utilize motion blur, which we refer to as a {\it 
    light flow,} to estimate depth.
%    revealed that it can be used 
%include the depth and to provide depth estimation algorithm from the flow.
Analysis reveals that minimum two {\it light flows}, which are retrieved from two 
    projected patterns on the object, are 
    required for depth estimation.
%To actually build the system, 
%two types of pattern are introduced, \ie,
%a sinusoidal pattern 
To retrieve two {\it light flows} at the same time,
%As for the actual pattern projection,
two sets of parallel line patterns are illuminated from two video projectors and 
    the size of motion blur of each line is precisely measured.
%By analyzing the apparent motion of the projected pattern, 
%{\ie, \it light flows}, object depth is estimated. 
By analyzing the {\it light flows}, \ie lengths of the blurs, scene depth information is estimated. 
In the experiments, 3D shapes of
    fast moving objects, which are inevitably captured with motion 
    blur, are successfully reconstructed by our technique.
\vspace{-2mm}
\end{abstract}

%%%%%%%%% BODY TEXT
\section{Introduction}
\vspace{-2mm}
3D shape reconstruction techniques by 
projecting special light to objects
(\ie, active lighting techniques)
have been important research topics. 
%This technique differs from shape reconstruction in natural lighting conditions, \ie,
%active lighting techniques. 
%Active lighting
%techniques 
Since they
have significant advantages, \eg, a textureless object can be robustly reconstructed
%, dense shapes are reconstructed 
with dense and high accuracy 
irrespective of light conditions, commercial 3D scanners are primarily based on active 
lighting techniques.
Among two important active lighting techniques, such as structured light and 
photometric stereo, %techniques,
structured light techniques, which 
encode the positional information of projector pixels into projected patterns, 
have been widely    
researched and developed due to their robustness, high precision, 
and density.

Recently, capturing 3D shapes of moving objects from a moving sensor 
has become
increasingly important, 
%. Capturing the 3D shapes of moving objects is required in 
%wide range of applications, 
such as measuring a scene from self-driving cars, etc., and solution is 
strongly desired.
%nursing, rehabilitation, and surveillance systems that scan human movement. %from world wide aging phenomena.
%Under the condition, it is impossible to capture a sequence of images of a static scene, which is 
%required for both common structured light and photometric stereo techniques.
% for reconstruction;
%thus, they cannot be used to capture a moving object 
%and/or from a moving sensor. 
The one-shot scanning technique, which is one of structured light techniques, 
is a promising technique because it required just single image and unaffected by motion.
%  that has been widely researched. % and developed. 
In general, one-shot scanning techniques embed the pattern's positional information into a small area
of the pattern for decoding~\cite{Kinect,Kawasaki:CVPR08,mesa,Ulusoy:3DIM09}, and thus, 
%To recover the shape from single image, 
the projected pattern tends to be complicated to increase robustness and density.
%and should be captured clearly for each frame. 
If object/sensor motion is faster than an assumption, the
reflected patterns are easily blurred out, resulting in unstable and inaccurate reconstruction.

To solve such motion blur problem, one may consider applying an algorithm 
developed for deblurring photographs. % as the solution. 
However, although the phenomena look similar, the optical processes are completely
different\footnote{Imagine the case where a planar board moves perpendicular 
direction to its surface normal. It is obvious that no blur effect appears on 
projected pattern.} and such algorithms cannot be applied to structured light.
%deblurring reflected patterns.
%
%it is impossible to simply apply the algorithm. 
In this paper, we analyze the information that is implied in the motion of 
a projected pattern (hereafter ``{\it 
light flow}'') to recover the shape. 
In fact, it is shown that 
{\it light flows} are explained  by three factors
such as depth, normal, and velocity of the object surface
and extracting one of them from a single {\it light flow} is impossible, but
possible with two {\it light flows}.
%In fact, 
%it is shown that light flow information simultaneously depends on the depth, normal, and 
%velocity of the object, 
%and extracting either of these values from a single flow information is not possible. 
In this paper, we propose a technique to decouple the three values from multiple
{\it light flows}. 
It is also revealed that {\it light flows} includes only little depth 
information, and thus, accurate detection of motion blur is required for 
practical depth estimation. To achieve this, we 
project the pattern of parallel lines onto the objects so that a blur size is precisely measured 
as the width of the band of motion blur of each line. % pattern. % s are and developed several special image processing techniques.
%As the projected pattern, we use images of  parallel lines. 
%From the captured image, the length of 
%blur can be %robustly 
%retrieved %from the observed image 
%as the width of the band of a
%blurred line. 
\fcut{
Because the length of the line interval can also be retrieved robustly, the 
flow ratio can be calculated by simply using two lengths and the ratio between 
two projectors can be calculated as a ratio of the two ratios.
}
%Motion blur is known to be observed in photograph where objects and/or a camera 
%move faster than shutter speed of the camera~\cite{}. The blur of reflected pattern is 
%different optical phenomina, however, 
%We offer the following contributions.
%\begin{enumerate}
%\item Depth from {\it light flows}, which does not require 
%%      pattern information, 
%      a decoding process, or a matching process, is presented.
%\item Shape recovery of ``fast motion,'' where motion speed is faster than the
%      shutter speed causing blurred patterns, is realized.
%\item Two algorithms using two different patterns are proposed to reconstruct a shape.
%\end{enumerate}

The contributions of this paper is as follows:
(1) Depth from {\it light flows}, which does not require 
%      pattern information, 
      a decoding process, or a matching process, is presented.
(2) Shape recovery of ``fast motion,'' where motion speed is faster than the
      shutter speed causing blurred patterns, is achieved.
(3) Two algorithms using two different patterns are proposed to reconstruct  
shapes of fast moving objects.
%Our technique can be used  for scanning moving objects and
% scanning an entire scene from a moving sensor, such as vehicle-mounted 3D sensor.
%In addition, the technique is based on a pixel-wise reconstruction.
%shape and velocity of 
%Therefore, 
By using our technique,
multiple fast moving objects, which are almost impossible to
scan with state of the art techniques, can be reconstructed, for example, %. We demonstrate that 
rotating blades of a fan are reconstructed by using a normal camera and a video 
projector. 

\section{Related works}

%3D shape reconstruction techniques using active light have been widely researched.
%Among them, structured light and 
%photometric stereo techniques are common.

%
There are two major shape recovery techniques using active light, such 
as photometric stereo and structured light.
Photometric stereo recovers the surface normal of each pixel
using multiple images captured by a camera while changing the light source 
direction~\cite{ikeuchi1981determining,Horn:1989:OSS:93871.93877,Higo2010,Maier-Hein2014}. 
%Since necessity of multiple images, the technique is not suitable to capture moving objects.
%In addition, 
Although photometric stereo %techniques 
%it 
can recover surface normals, %but 
they cannot recover absolute 3D distances;
%shape, 
thus, their applicability is limited.
The structured light technique has been used for practical applications~\cite{Salvi:PR04,DBLP:conf/eccv/WangSGN16,O'Toole:2015:HCE:2809654.2766897}.
% in many areas 
%due to its robustness, accuracy, and 
%practicality
 Two primary approaches to encode positional information into patterns
are temporal and spatial encoding. 
%There are two primary approaches to 
%embedding the positional information of a projector pixel into a pattern, such as 
%temporal and spatial encoding. 
Because temporal encoding requires 
multiple images, it is not suitable for capturing moving 
objects~\cite{sato1985three,
taguchi2012motion}. 
%Conversely,
%spatial encoding, which is also known as the one-shot scanning technique, 
Spatial encoding
requires only a single image
%; therefore, it 
and 
is possible to capture fast-moving objects~\cite{Kinect,Kawasaki:CVPR08,mesa,Ulusoy:3DIM09}. For example, 
some commercial devices realize 30 fps at VGA resolution~\cite{Kinect}. Another 
technique can scan the shape of a rotating blade~\cite{Kawasaki:CVPR08}.
Although such techniques can capture fast moving objects, they
still require 
%a pattern to be decoded for 
%shape reconstruction. Thus, 
%the pattern must be captured clearly. 
the patterns to be captured clearly. 
%Otherwise,
%the pattern becomes blurred due to unexpected fast motion, and the accuracy 
%is reduced or the entire shape cannot be reconstructed. 
If the pattern is blurred due to unexpected fast motion, 
the accuracy is reduced or reconstruction fails.
Kawasaki \etal needed to use 
a high-speed camera
to measure a rotating blade~\cite{Kawasaki:CVPR08}.
%In fact,  
%a rotating blade was reconstructed with a high-speed camera and 
%high-performance 
%video projector~\cite{Kawasaki:CVPR08}. 
In contrast, our technique uses a 
different approach, \ie, shapes are reconstructed from 
motion
%, typically,  
blur of a projected pattern. This helps to decrease the 
difficulty for scanning fast moving objects and allows off-the-shelf devices to 
%work with 
be used for 
such extreme conditions. 
%The proposed technique was used to 
%successfully reconstructed rotating blades using a normal camera and projector.

Another limitation of structured light techniques is that they are highly
dependent 
%on the decoding ability of the positional information. 
on the locally decoded pattern information. 
If decoding the positional information fails for some 
reason, such as noise, specularity, blur, etc., shape reconstruction will 
subsequently fail. 
%In addition, if pattern information is not known, shapes 
%cannot be reconstructed at all. 
To avoid this limitation, 
%there are techniques 
%that 
some techniques
are based on geometric constraints rather than
decoding~\cite{Sagawa:ICCV09,proesmans1997one,koninckx2006real,Ulusoy:3DIM09} or 
active usage of defocus effect by coded aperture~\cite{kawasaki2015active}.
One problem with such techniques is 
that, because they are heavily dependent on geometric constraints or special devices,
strong motion blur cannot be handled.
%pattern detection errors will not only gradually degrade the distance measure but also 
%leads to total reconstruction failures. 

\begin{figure}[t]
%\vspace{-5mm}
\begin{center}
	\includegraphics[width=60mm]{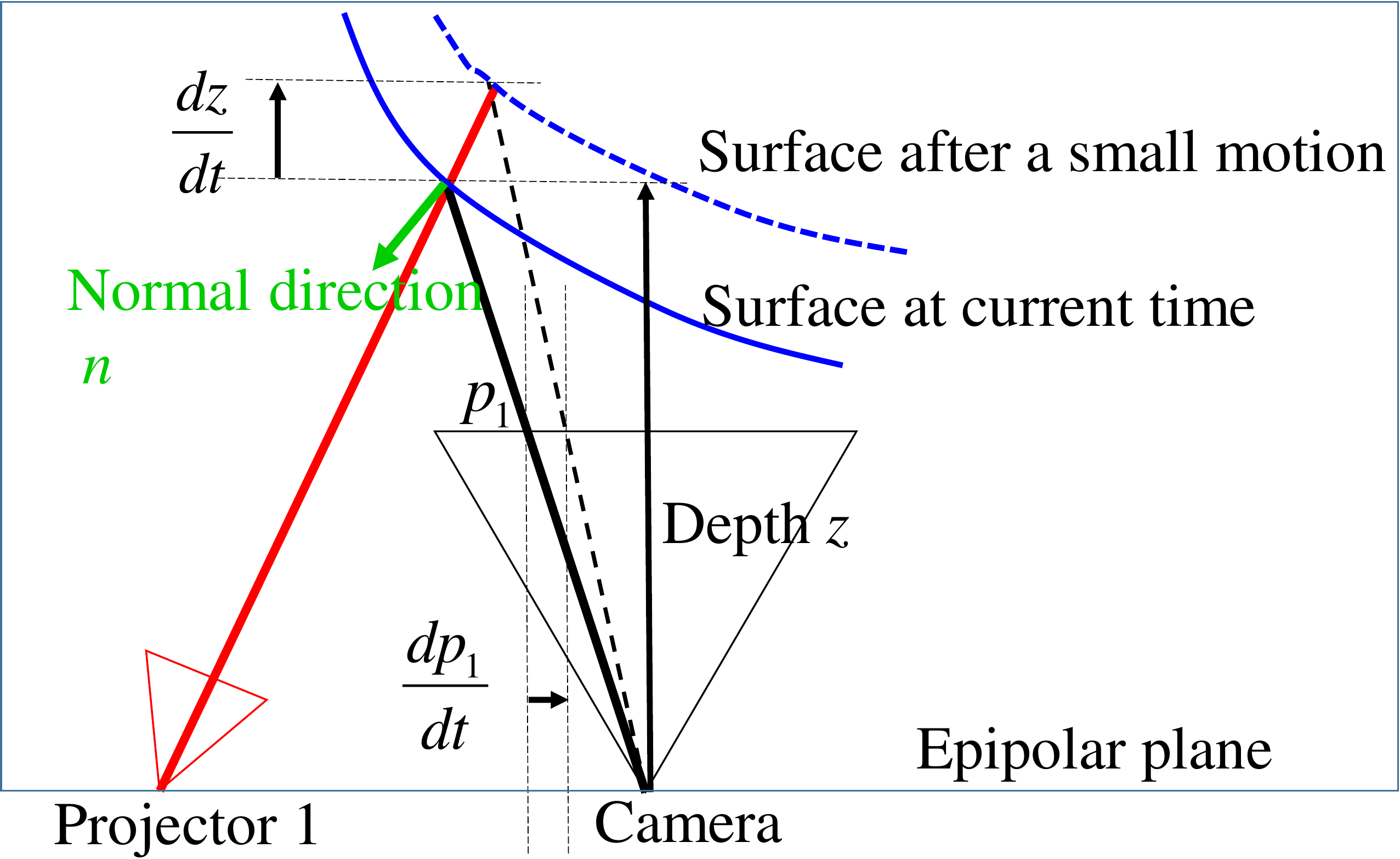}
	\caption{Motion of the object surfaces and patterns. Note that his figure represents a 2D section of the scene by an epipolar plane.}
	\label{fig:overviewsymbols}
 	\vspace{-5mm}
\end{center}
\end{figure}

%Our 
%technique is also decode free, and thus, robust to noise, blur and other turbulences.
Our technique uses relationship between two overlapped patterns and such 
techniques have been researched.
%There are shape 
%reconstruction methods where relationship of overlapped two or more patterns are 
%used. %investigated.
%Another related area is shape recovery from patterns overlapping. 
The Moire method 
is a well-known technique where a high-frequency pattern is projected on an
objected and is observed by a high-frequency gate, which generates Moire patterns 
%at 
%the same height~
\cite{Takasaki:73,han2001moire}. Recently, a technique that uses 
the intersection of 
multiple parallel line patterns was proposed to achieve an entire scene 
reconstruction~\cite{Furukawa:PSIVT2010}. Both techniques assume a static scene and recover 
shape from geometric constraints, whereas our technique recovers shape from 
a temporal gradient. %such as flow of pattern, 
To the best of our knowledge, no such technique has been proposed previously.

\section{Analysis of projected pattern flow}
%節タイトル：analysis of projected pattern flowぐらいのほうが適切？
\subsection{Overview}
\label{sec:overview}

In this section, we briefly overview information that can be obtained 
from apparent motion of projected patterns. 
%projected onto the object surfaces
%for a projector-camera system.  
We assume that the projectors and the camera are relatively static and calibrated. 
If the object moves, the observed patterns move along epipolar lines. 
\fnoteIII{
For the target scene, we assume that the regions around the measured 3D points 
can be regarded as locally planer and depth of the points are changing. 
Otherwise, we would not be able to observe continuous motions of patterns
moving with the scene motions. 
}

Without losing generality, the relationships between the motions of the target surfaces and the patterns 
can be considered within the epipolar planes. 
Fig.~\ref{fig:overviewsymbols} shows the relationship. 
%ここでは、パターンプロジェクタから物体表面に投影されるパターンの、カメラから見た時の見かけの動きから、
%物体のgeomtryと動きについて、どのような情報が得られるのか、解析する。
%パターンプロジェクタとカメラはstaticであり、calibration済とする。
%物体が動く場合、カメラから観察されるパターンは、エピポーラ線に沿って動く(light flow)。
%この時のパターンの動きの性質は、
%一般性を失うことなく、
%エピポーラ平面内で考えてよい。図。

Let the apparent position of the pattern be $p$, and the apparent motion of the pattern (\ie, light flow) be $\frac{dp}{dt}$.
$\frac{dp}{dt}$ depends on the surface depth $z$, surface normal $n$, and the depth velocity $\frac{dz}{dt}$
as shown in Fig.~\ref{fig:overviewsymbols}.
$\frac{dp}{dt}$ is proportional to $\frac{dz}{dt}$, and represented as
\begin{equation} 
\frac{dp}{dt} = F(z,n) \frac{dz}{dt},
\label{eqn:floweq}
\end{equation} 
where $F$ is a nonlinear function that can be defined from the projection model and calibration parameters of the projectors and cameras
\fnoteIII{
(\ie, the information of the epipolar geometry of the projector-camera pair
is included in $F$).
%Since 
%Fig.~\ref{fig:overviewsymbols} 
By limiting the geometry within a epipolar plane,
 the normal direction at a point on the surface
 can be represented by a scalar variable, thus, 
$F(z,n)$ is a 2D real function. 
}

%パターンの見かけの位置を$p$とし、パターンの見かけの速度を$\frac{dp}{dt}$とすると、
%$\frac{dp}{dt}$は、物体表面の奥行き$z$、法線方向$n$、物体の奥行きの変化率$\frac{dz}{dt}$(カメラの光軸方向への速度）によって決まる。
%このとき、$\frac{dp}{dt}$は物体表面の速度 $\frac{dz}{dt}$に比例し、
%\begin{equation} 
%\frac{dp}{dt} = F(z,n) \frac{dz}{dt}
%\end{equation} 

%となる。ここで、$F$はキャリブレーションパラメータから幾何学的に定義可能な非線形関数である。
%エピポーラ平面内に限定すると、$n$は1自由度なので、$F(z,n)$は２次元実数関数である。

%ここで、パターンが$N$あるとする。それぞれのパターンについて
%light flowを計測し、$z$及び$n$を推定することを考える。

Here, we assume that we have $N$ projectors. 
The light flows can be observed for each projector.

%\subsubsection{Case of $N=3$}

%\subsubsection{$N$=3の場合}
In case of $N=3$, where we use three patterns from three projectors, 
the observed light flows are 
$\frac{dp_i}{dt}$ where $i=1,2,3$, and we obtain three equations
\begin{align} 
\frac{dp_i}{dt} & = F_i(z,n) \frac{dz}{dt}, (i=1,2,3)
\end{align}
where $F_i$ is the same function with Eqn.~\ref{eqn:floweq} for
pattern $i$.
%\frac{dp_2}{dt} & = F_2(z,n) \frac{dz}{dt}\\
%\frac{dp_3}{dt} & = F_3(z,n) \frac{dz}{dt}
%\end{align}
%３個のプロジェクタを利用する場合、
%$\frac{dp_1}{dt}, \frac{dp_2}{dt}, \frac{dp_3}{dt}$ の３つの値が観測できる。
%\begin{align} 
%\frac{dp_1}{dt} & = F_1(z,n) \frac{dz}{dt}\\
%\frac{dp_2}{dt} & = F_2(z,n) \frac{dz}{dt}\\
%\frac{dp_3}{dt} & = F_3(z,n) \frac{dz}{dt}
%\end{align}
%の3式が利用できる。
%$\frac{dz}{dt}$を消去すると、
%\begin{align} 
%\frac{dp_1}{dt} / \frac{dp_2}{dt} & = F_1(z,n) / F_2(z,n)\nonumber \\
%\frac{dp_2}{dt} / \frac{dp_3}{dt} & = F_2(z,n) / F_3(z,n)
%\label{eqn:flowratio}
%\end{align}
%の関係式が得られる。
By eliminating $\frac{dz}{dt}$, we obtain two equations:
\begin{align} 
\frac{dp_1}{dt} / \frac{dp_2}{dt} & = F_1(z,n) / F_2(z,n)\nonumber \\
\frac{dp_2}{dt} / \frac{dp_3}{dt} & = F_2(z,n) / F_3(z,n).
\label{eqn:flowratio}
\end{align}
%これらの式は、未知数が$z$,$n$の２変数であり、$F_1,F_2,F_3$は
%既知関数である。2個の制約式があるので、
%非線形方程式の求解法によｒ、$z$と$n$を求めることが可能である。
The unknown variables are $z$ and $n$, and $F_1,F_2,F_3$ are known 
functions since they are defined 
from calibration parameters. 
Since we have two equations with two unknowns, 
we can estimate $z$ and $n$ by numerically solving these equations.

%また、\ref{eqn:flowratio}より、
%幾何学的な情報は、パターンの速度$\frac{dp_i}{dt}$そのものではなく、
%異なるパターンの速度の比$\frac{dp_i}{dt}/\frac{dp_j}{dt} (i\ne j)$
%から得られることがわかる。

Note that, from Eqn.~\ref{eqn:flowratio}, 
geometrical information are obtained from ``ratio'' of the pattern motions
 $\frac{dp_i}{dt}/\frac{dp_j}{dt} (i\ne j)$, rather than 
 $\frac{dp_i}{dt}$ themselves.

%\subsubsection{Case of $N=2$}
In case of $N=2$, we obtain only one equation by eliminating  $\frac{dz}{dt}$, 
which is
$\frac{dp_1}{dt} / \frac{dp_2}{dt}  = F_1(z,n) / F_2(z,n)$.
Since we have one equation with two unknowns, 
we cannot determine $z,n$ for this case.

%の2式から
%$\frac{dp_1}{dt} / \frac{dp_2}{dt}  = F_1(z,n) / F_2(z,n)$
%が得られるが、未知数が2個で、等式が1個なので、これらの式のみからはz,nを求めることはできない。

%ただし、次節で述べるように、投影パターンとしてunoiformly spaced parrarrel patternなどを
%仮定すれば、パターンの動きと、平行線の見かけのintervalの情報から、２個のプロジェクタから２個の
%パターンを投影することで、zを推定することが可能である。
%さらに、N=2の特殊な場合として、２個のパターンを、単一のプロジェクタから投影することで、
%プロジェクタ１台、カメラ１台での計測を行うこともできる。

However, if we use additional information such as knowledges of the projected patterns, 
we can estimate $z$.
As described in the following sections, 
we propose to use uniformly spaced parallel patterns from ``two'' projectors 
to estimate depths. 
Moreover, we also show a special case of $N=2$, 
where the two patterns are projected from a ``single'' projector. 
In the case of single projector, 
we should use non-uniformly spaced parallel pattens. 

%\subsubsection{Case of $N=1$}
%
%\subsubsection{N=1の場合}
%単一のパターンの運動を観測した場合、
%$\frac{dp_1}{dt}  = F_1(z,n) \frac{dz}{dt}$,
%の単一の式が得られるが、この式のみからは、$z, n, \frac{dz}{dt}$いずれの情報も確定できない。
In case of $N=1$, we use only one pattern,
we obtain only
$\frac{dp_1}{dt}  = F_1(z,n) \frac{dz}{dt}$.
Obviously, we cannot solve $z, n, \frac{dz}{dt}$ from the form. 

As explained here, we can estimate depths if we can observe three patterns from three projectors.
However, to achieve this, three patterns should be decoupled from captured images,
which is not an easy task. 
One solution may be using three color channels, however, crosstalks between color channels are problematic.
%large problems. Especially, 
%crosstalks between red-green channels and blue-green channels are sometimes large.
%このように、３個のプロジェクタを利用して、patternの動きを取得すれば、物体の奥行きを
%求めることが可能であるが、このためには、３個のプロジェクタのパターンを、
%キャプチャ画像から互いに分離する必要がある。
%このための方法としては、色チャンネルを使うことが考えられるが、３色を分離する場合、
%色チャンネル間のクロストークが大きな問題となる。特に、
%ＲＧ間と、ＢＧ間のクロストークは大きい。
On the other hand, decoupling two patterns are relatively stable, since crosstalks between red-blue channels are small. 
Thus, in this paper, we propose a technique which requires only two colors, \ie 
two projection patterns, for shape reconstruction. 

\subsection{Depth estimation with two-pattern case} 
\label{sec:depthfrom2proj}

%●推定について
%
%・法線の影響を受ける→2自由度ある
%・法線推定→空間微分dxとdyは法線成分を持っているので、これを使用？（SfShadin
%g？）
%
%Unknown
%・法線（シータ、ファイ）、Disparity、Disparityの変化量→4つ
%入力
%・時間差分*2、空間差分*2→4つ
%
%拘束式と未知数の数はあっている→解けるのでは？
%
%順方向モデルを作成中→Disparityの変化に対するセンシティビティ調査が必要
%逆方向はその後考える

%物体をカメラから観測するときに、特定の画素$p$について考える。
%$p$に対応するカメラからのrayを$\bf r$とし、
%物体と$\bf r$の交点を$\bf s$とする。
%カメラから見た$\bf s$の奥行きを$d$とする。
%このとき、
%\[
%{\bf s} = d~{\bf r}
%\]
%である。
%物体がカメラに対して移動しているとき、
%カメラからの奥行き$d$が変化する。
%$d$の微小な変化を
%$\Delta d$とする。
%このとき、
%$\bf s$の位置は、
%\[
%{\bf s}+ {\Delta }{\bf s} = (d + \Delta d){\bf r}
%\]
%に変化する。

%\ref{sec:overview}
%で述べたように、見かけのパターンの動きのみからdepth情報を求めるには、
%３個のパターンを観測する必要がある。
%しかし、パターンについての知識を利用することで、より少ないパターンの動きから、
%depthを求められることが可能である。
%本節では、平行線のパターンを仮定し、平行線の見かけ上の間隔を利用することで、
%２個のパターンの観測結果からdepth情報を推定する。

%As previously described, estimation of depths from only the apparent motion of the pattern
%needs three patterns from three projectors. 
%However, by using knowledge of patterns,
%we can estimate depth from less pattens. 

In this section, 
we explain depth estimation with two patterns using uniformly spaced parallel lines.
%The approach to achieve this is estimating the {\it light flows}
To achieve this, we estimate the {\it light flows}
%by the projectors' image coordinates, 
``in the projectors' image coordinates''
instead of 
%by the camera's image coordinate. 
in the camera's image coordinates. 
Then, we can avoid considering normals of the surfaces 
as shown in the the following discussions.
%by the following discussions.

\begin{figure}[tb]
%\vspace{-5mm}
\begin{center}
	\includegraphics[width=0.36\textwidth]{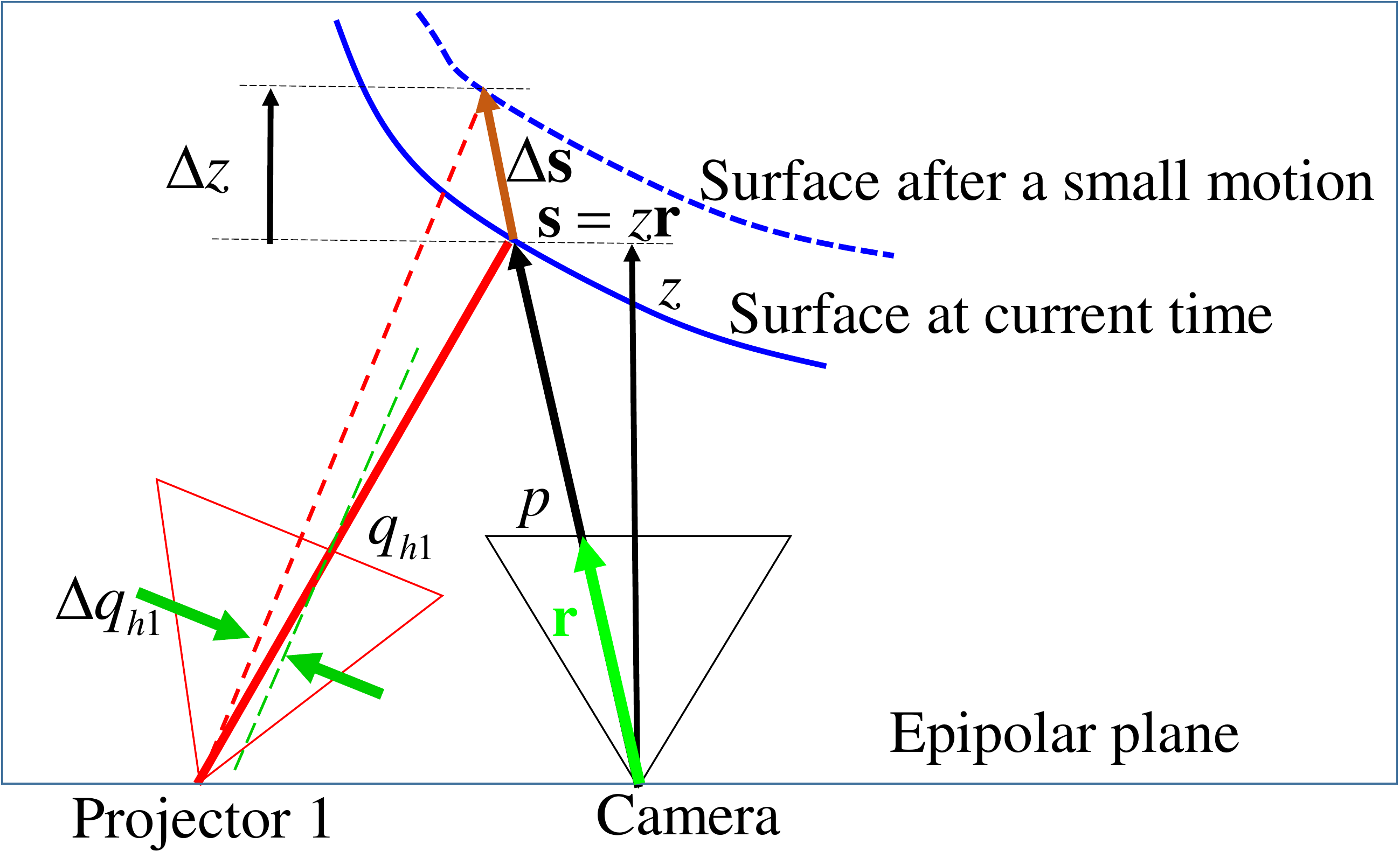}
% 	\vspace{5cm}
	\caption{Light flows observed with respect to the pattern coordinates $q_h$.}
	\label{fig:symbols2proj}
\vspace{-5mm}
\end{center}
\end{figure}

%\snote{r,q,sについての図が欲しい}
First, we assume a pixel $p$ while observing a target object from a camera. 
We define the ray from the camera corresponding to $p$ as $\bf r$,
the intersection between the object and $\bf r$ as $\bf s$,
and the depth of $\bf s$ as $z$ . Thus, 
\begin{equation} 
{\bf s} = z~{\bf r}.
\end{equation} 
We assume that the object is moving with respect to the camera;
thus, depth $z$ changes.
Let a small displacement of $z$ be $\Delta z$.
Then, 
the position of $\bf s$ is expressed as follows:
\begin{equation} 
{\bf s}+ {\Delta }{\bf s} = (z + \Delta z){\bf r}.
\end{equation} 
Fig. \ref{fig:symbols2proj} shows the relationships between the symbols
used in this section. 

\begin{figure}[tb]
%\vspace{-5mm}
\begin{center}
	\includegraphics[width=0.47\textwidth]{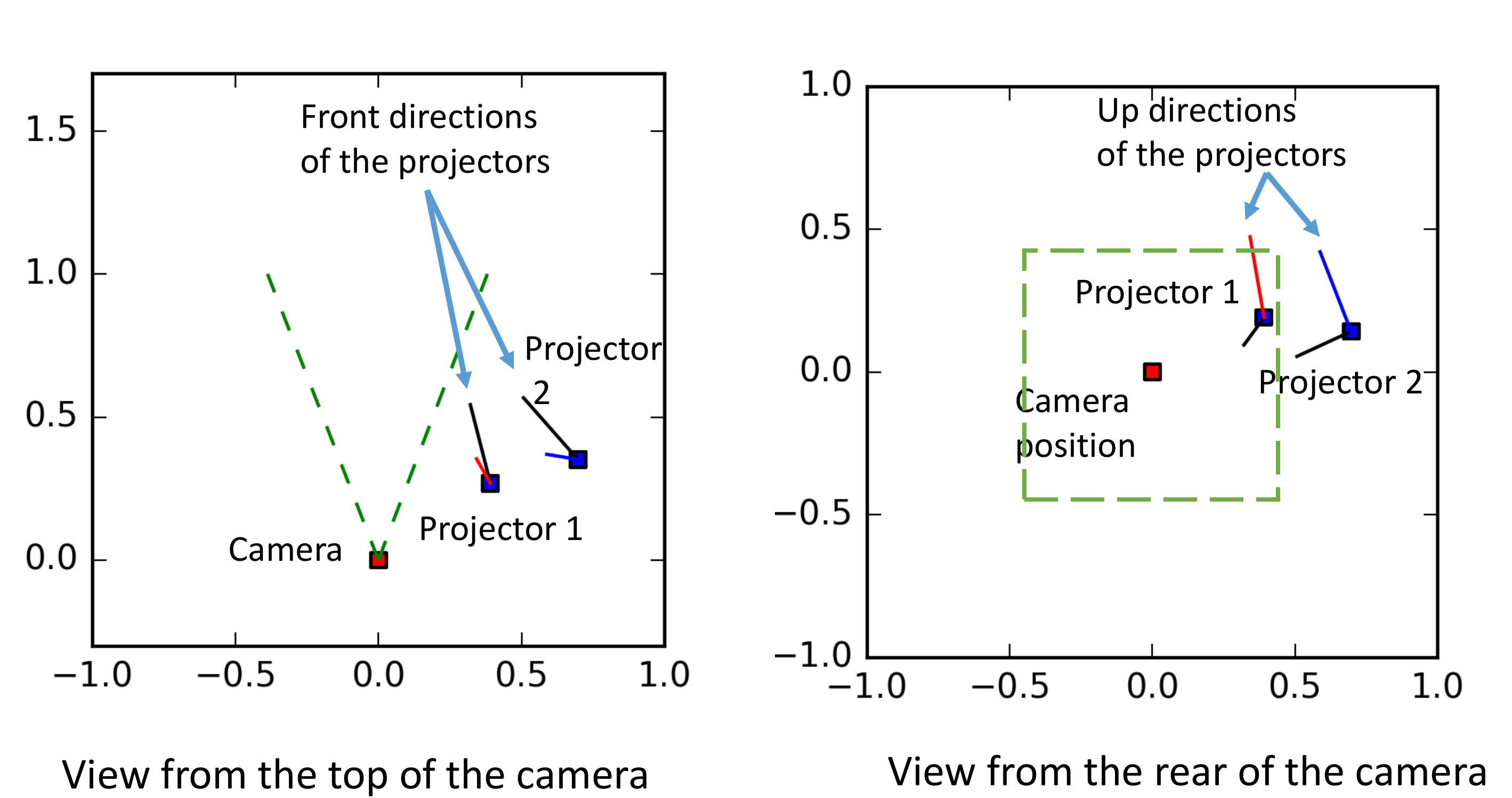}
% 	\vspace{5cm}
	\caption{Positions of the camera and the projectors.}
	\label{fig:setup}
\vspace{-5mm}
\end{center}
\end{figure}

\begin{figure}[tb]
%\vspace{-5mm}
\begin{center}
	\includegraphics[width=55mm]{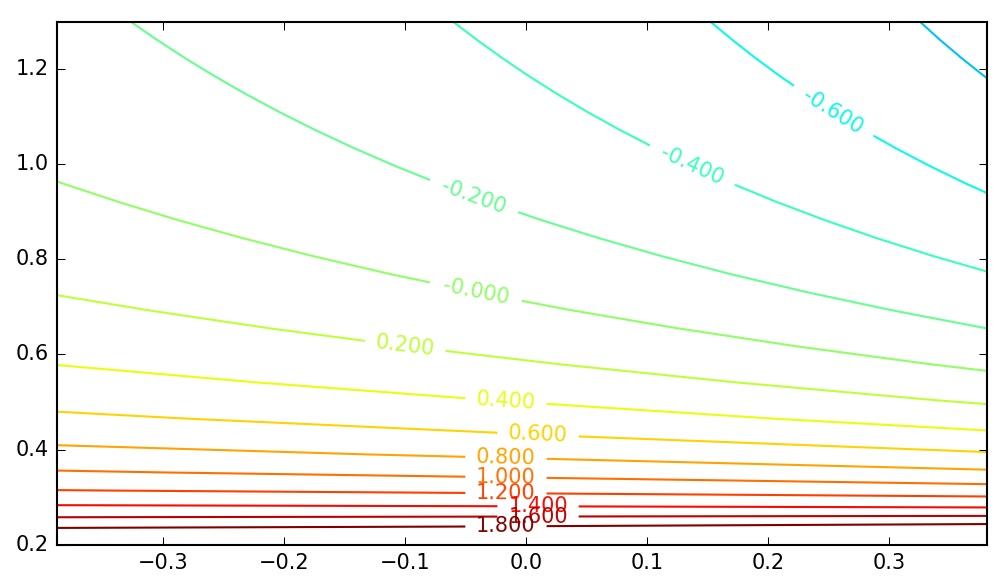}\\
	(a): $h(z)$ for the setup shown in Fig.~\ref{fig:setup}.\\
	\includegraphics[width=65mm]{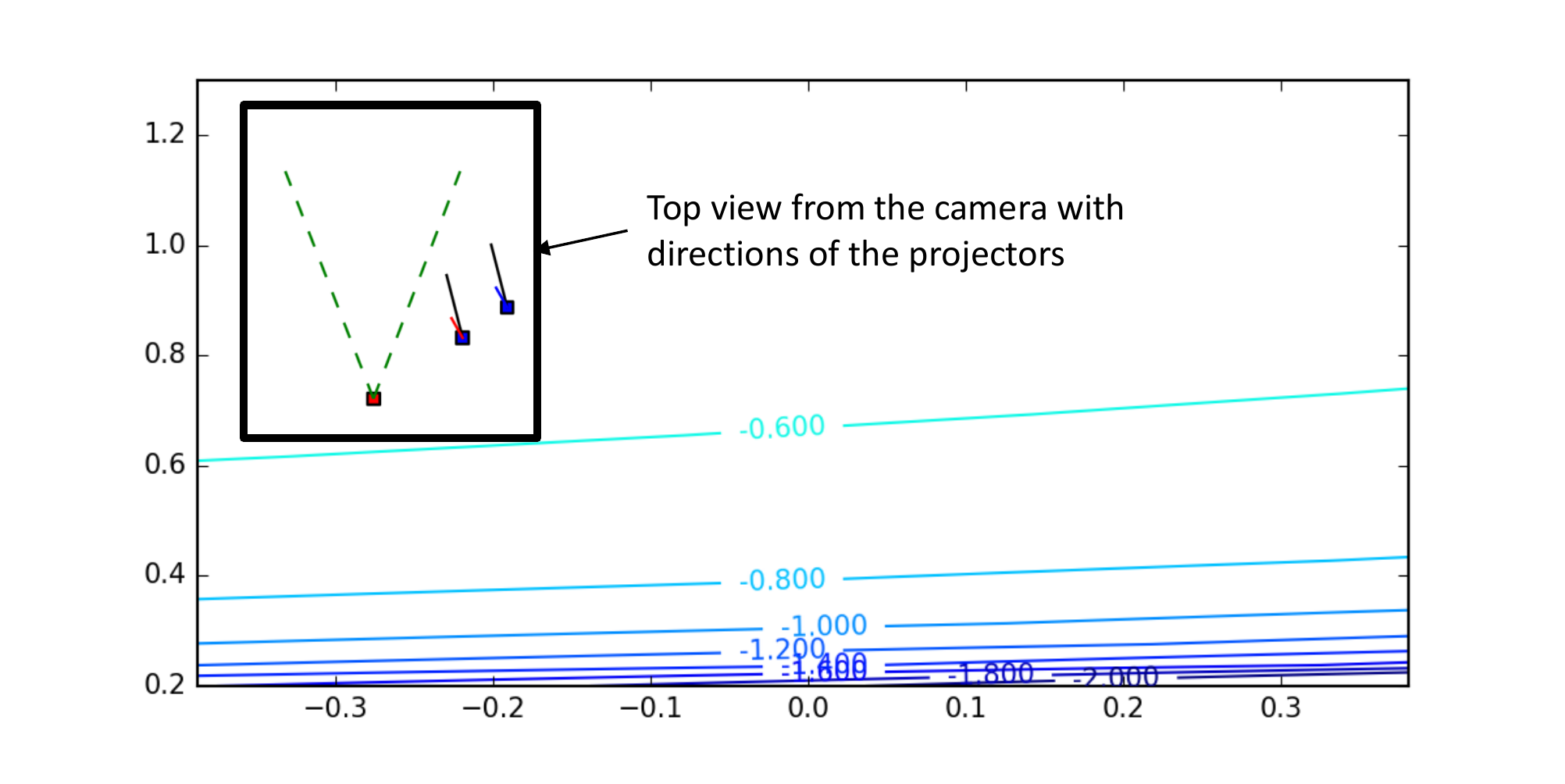}\\
	(b): Bad setup where gradient of $h(z)$ becomes small.\vspace*{3mm}
	\caption{Function from the depth 
%	from the camera 
	to the log ratio of {\it light flows}
%	pattern displacements of the two projectors
	( $h(z)$ of Eqn.~\ref{eqn:depthflowratiofunc}) :
	(a) The experimental setup used in this paper.
	The horizontal axis is the $x$ position of the camera in the coordinates of the normalized camera.
	(b) A bad setup where projectors 1 and 2 are placed in parallel. }
%	\knote{センシティビティの図}}
	\label{fig:depthfigratiofunc}
\vspace{-5mm}
\end{center}
\end{figure}
%$s$の位置の変化に伴い、
%$s$を投影するプロジェクタパターンが変化する。

The projector can be geometrically regarded as an ``inverse camera,''
\ie, the relationship between a 3D coordinate ${\bf s}$ of point $s$
and the 2D coordinate on the pattern pixel coordinate ${\bf q}$
that illuminates $s$ can be formulated in the same as the projections of the 
camera. This can be expressed as follows:
\begin{equation}
{\bf q}={\bf f}({\bf R}{\bf s} +{\bf t}),
\end{equation}
where ${\bf s}$ is the coordinates of $s$ in the camera coordinate system;
${\bf R}$ and ${\bf t}$ are the rotation matrix and the translation vector of the rigid transformation from the camera coordinates 
to the projector coordinates, respectively; and
${\bf f}$ is a perspective projection function 
that is the same as the normalized camera.
We assume that the 
pattern image illuminated from the projector is 
%constant in the vertical axis. 
parallel lines that is near the vertical  direction, 
and the epipolar lines are assumed to be near the horizontal direction. 
%垂直に近い平行線のパターンであり、
%エピポーラ線は水平方向に近いものとする。
Thus, 
the pattern intensity only varies with the 
horizontal coordinate, 
which we define as $q_h$.
Then,
\begin{equation}
q_h=f_h({\bf R}{\bf s} +{\bf t}) 
=f_h(z{\bf R} {\bf r} +{\bf t}), 
\end{equation}
where $f_h( (x,y,z)^{\mathrm{T}} )\equiv (-x/z)$.

%The motion of the light pattern, \ie, {\it light flow}, is observed as the 
%changes of the pattern coordinates $q_h$.
%投影パターンが、等間隔の平行線であり、
%各パターンの動きと、パターン同士の間隔が観測できる場合、
%パターン同士の間隔に対するパターンの動きの大きさの比率によって、
%light flowの大きさを、パターン画像上でのパターンの座標
%$q_h$の変化として、表すことができる。

Let the relationship between the depth $z$ and $q_h$ 
be substituted by a function $g$. 
Thus, 
\begin{equation}
q_h=f_h(z{\bf R} {\bf r} +{\bf t})=g(z).\label{eqn:extrinsic}
\end{equation}
Here assume that
pattern coordinate 
$q_h$ is changed by $\Delta q_h$
for a small displacement $\Delta z$ of $z$.
$\Delta q_h$ can be observed
as
%(usingとするべきか)
{\it light flow}.
Then, 
the ratio between the changes 
can be approximated by the derivative of $g$:
\begin{equation} \label{eqn:deffarential}
\frac{\Delta q_h}{\Delta z} \approx g'(z).
\end{equation}
Function $g'(z)$
changes depending on the depth $z$.

Since two projectors are assumed in this analysis,
there are two pattern coordinates for each projector, such as 
 $q_{h1}$ and $q_{h2}$,
and two functions for each projector, such as
$g_1$ and $g_2$. Thus,
\begin{equation}
\frac{\Delta q_{h1}}{\Delta z} \approx {g_1}'(z), 
\frac{\Delta q_{h2}}{\Delta z} \approx {g_2}'(z)
\end{equation}
can be reduced to
\begin{equation}
\frac{\Delta q_{h1}}{\Delta q_{h2}} \approx \frac{{g_1}'(z)}{{g_2}'(z)}.
\end{equation}
Here, 
function $h$ is defined as follows:
\begin{equation}
h(z) \equiv \log \left ( \frac{{g_1}'(z)}{{g_2}'(z)} \right),
\label{eqn:depthflowratiofunc}
\end{equation}
where $h(z)$ is a function of $z$ and does not depend on $\Delta z$.
If $h(z)$ is a monotonic function in the domain of the working distances of $z$, 
there is an inverse function $h^{-1}$ for this domain,
and depth $z$ can be estimated 
from the ratio of the 
%displacements of the two patterns
two {\it light flows}
($\Delta q_{h1} $ and $\Delta q_{h2}$) as follows:
\begin{equation}
z \approx h^{-1}\left ( \log \left ( \frac{\Delta q_{h1}}{\Delta q_{h2}} \right) \right).
\end{equation}

Function $h(z)$ depends on the pixel position $p$,
\ie, for different pixel positions, $h$ becomes different functions.
Thus, let $h(z)$ for a pixel position $p$ be expressed as
$h_p(z)$.
\fnoteIII{
Similarly to $F$, the information of the epipolar geometry 
is included in $h$,
where the explicit dependency is based on Eqn.~\ref{eqn:extrinsic}.}

An example of the camera setup is shown in Fig.~\ref{fig:setup} 
(this setup is also used in the later experiments),
and $h_p(z)$ for this example setup
is shown in Fig.~\ref{fig:depthfigratiofunc}(a),
where $p$ is on a horizontal line in a camera image (including the image center)
%The horizontal axis is the position of $p$ and
and the vertical axis represents $z$.
As can be seen,
$h(z)$ actually varies depending on $z$
and 
$h_p^{-1}$ can be defined for each pixel.
In real implementation, 
we can sample pairs of 
$z$ and the function value $h_p(z)$
as 
$\{(z, h_p(z)) | z \in \{z_1, z_2, \cdots, z_n\} \}$
and 
$h_p^{-1}$ can be approximated by
interpolation of the samples.
Then, depth estimation can be processed efficiently.  

Note that the positional setup of the projector and the camera affects severely 
to the sensitivity of the depth estimation. 
For example, an example of a bad setup is shown
in Fig.~\ref{fig:depthfigratiofunc}(b),
where the two projectors are placed in parallel.
Although this setup look similar to the setup of Fig.~\ref{fig:setup},
the gradient of $h_p(z)$ becomes much smaller, 
thus, depth estimation using $h_p^{-1}$ becomes much sensitive to observation noises. 
Specifically, if the projectors and the camera are placed in perfect fronto-parallel configuration, 
$h_p(z)$ becomes constant and $h_p^{-1}$ does not exist.

In the proposed method, 
the required input value is
the ratio of the {\it light flows}
%displacements of the pattern coordinates
$\Delta q_{h1}/\Delta q_{h2}$, 
whereas  the absolute values of 
$q_{h1}$ or $ q_{h2}$ are not required.
If the pattern image is repetitive, 
the depth can be estimated 
from only the relative local changes as described later.
This means that we do not need to encode the absolute pixel position 
into the pattern image and this is an important advantage of our method.

The {\it light flows} are  
observed as the motion on the image planes of the projectors
in Fig.~\ref{fig:symbols2proj},
rather than the motion on the image plane of the camera as shown in  Fig.~\ref{fig:overviewsymbols}. 
Observing {\it light flows} on the projectors' image planes 
is possible if we use knowledge of the projected pattern, 
such as uniformly spaced parallel lines. 
As described later, we use 
the uniform intervals as ``scales'' on the projectors' image planes.
Also, in the discussion of this section,
normal directions of the object surface are not considered.
This is because the relationships between the light flows and the object motion
are considered on the fixed ``ray'' from the camera along ${\bf r}$.
%This simplification was possible 
%because we fix the camera's image plane coordinate
%by considering {\it light flows} in the projectors' image planes. 

%また、本節の議論では、物体表面の法線ベクトルは出てこなかった。
%これは、light flowを、
%カメラでの見かけのパターン運動ではなく、
%パターン画像上での座標で表したことにより、カメラの特定のrayでの議論に限定したためである。
%これは、パターンの間隔情報を使うことで、
%物体の法線情報を、implicitに利用したことに等しい。
%
$h_p(z)$ does not depend on 
the intrinsic parameters
of the projector-camera pair,
since 
the displacements 
$\Delta q_{h1}$ and $\Delta q_{h2}$
are represented in normalized coordinates.
%extrinsic parameters of the projector-camera pair,
%%the positions and directions of the projectors 
%rather than
%the intrinsic parameters.
% of the projector.
%However, 
%the displacements of normalized coordinates 
%$\Delta q_{h1}$ and $\Delta q_{h2}$
%are not normally observable,
%and the 
%displacements in the units of pixels
%$\Delta q_{hp1}$ and $\Delta q_{hp2}$
%are observed instead.
For real measurements, 
displacements in the units of projector pixels 
$\Delta q_{hp1}$ and $\Delta q_{hp2}$
are observed instead.
%Thus, 
The ratio of these values can be converted by 
%Here let the focal length of the projectors 
%be $f_1$ and $f_2$. Then, we obtain the following:
\begin{equation}
\frac{\Delta q_{h1}}{\Delta q_{h2}}
=
\frac{f_2}{f_1}
\frac{\Delta q_{hp1}}{\Delta q_{hp2}},
\end{equation}
where 
$f_1$ and $f_2$ are the focal lengths of the projectors.

\begin{figure*}[tb]
%\vspace{-5mm}
\begin{center}
	\includegraphics[width=26mm]{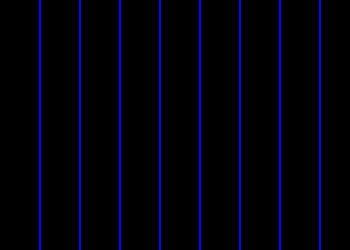}
	\includegraphics[width=26mm]{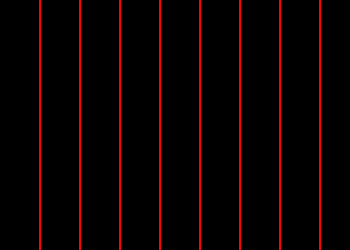}
	\includegraphics[width=26mm]{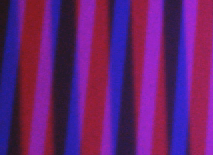}
	\includegraphics[width=26mm]{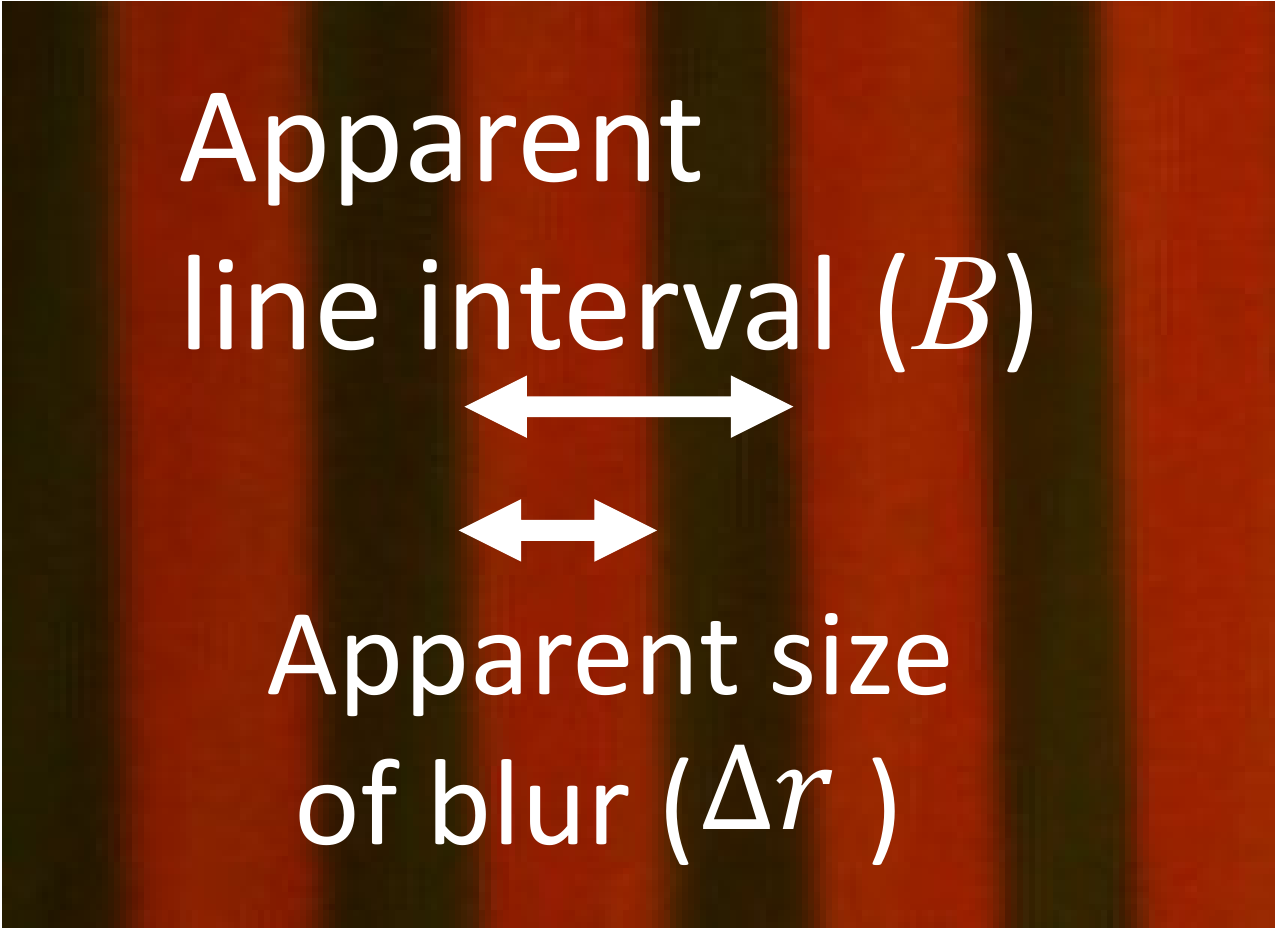}
	\includegraphics[width=54mm]{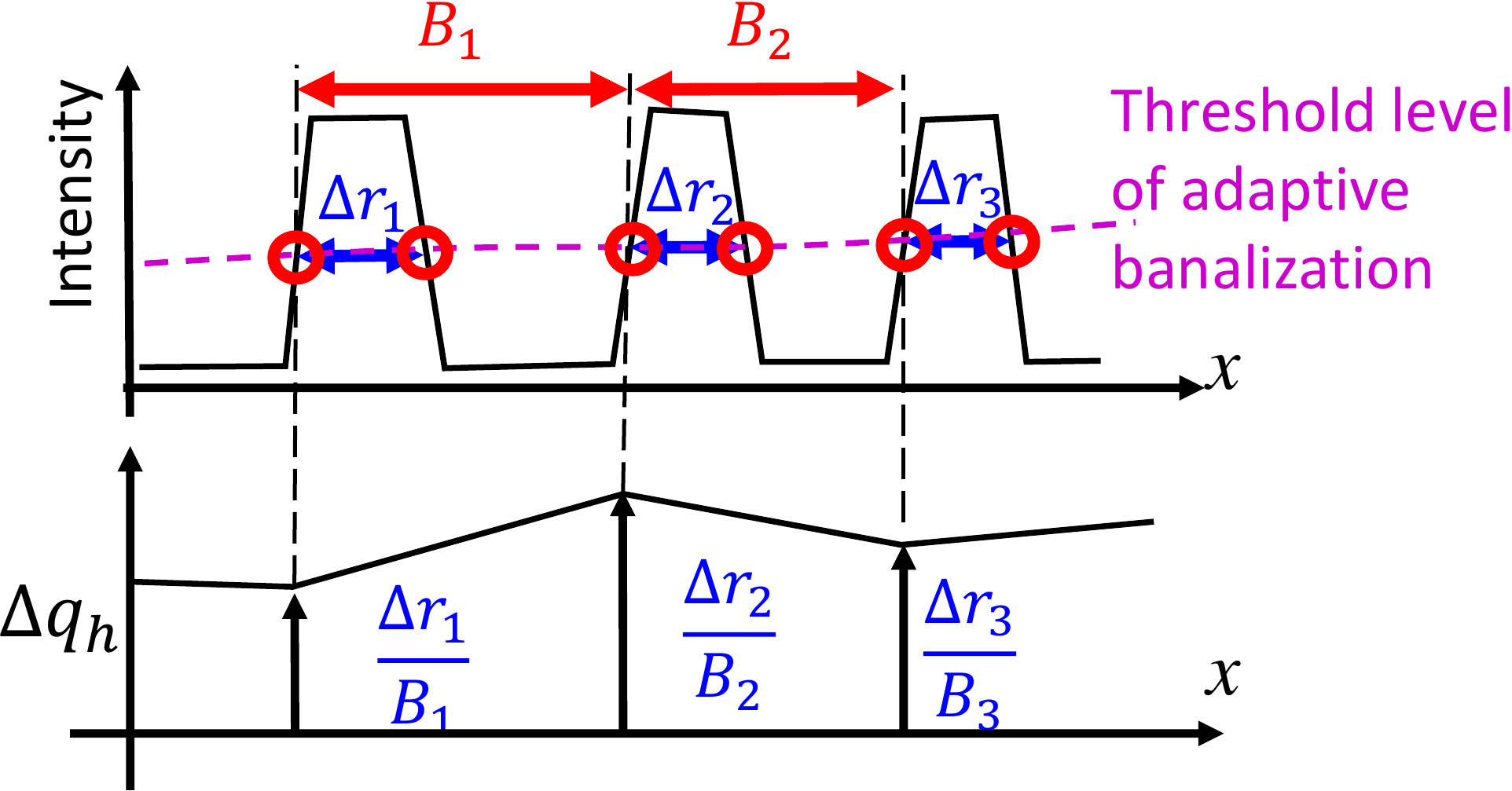}\\
	\hspace*{-20mm}
	(a)\hspace{22mm}
	(b)\hspace{22mm}
	(c)\hspace{22mm}
	(d)\hspace{38mm}
	(e)
%	\includegraphics[width=65mm]{figure/Equipment_OpticalSystem3.eps}\\
%	\includegraphics[width=35mm]{fig/ball-with-lines.png}
%	\includegraphics[width=35mm]{fig/ballindark.png}
%	\includegraphics[width=35mm]{fig/band0.png}
%	\includegraphics[width=35mm]{fig/lineflow-crop.pdf}
% 	\vspace{10cm}
	\caption{{\it Light flow} estimation from uniform line patterns: (a,b) line pattens,
	(c) projected line patterns with motion blurring,
	(d) an apparent line interval and a size of blur,
	(e) flow estimation using line intervals and the blur sizes.}
	\label{fig:flowestimate-line}
	\vspace{-3mm}
\end{center}
\end{figure*}

%\section{Implementation}
%\section{Experimental systems and light flow estimation}
\section{Experimental systems}

\subsection{System configurations} %\knote{川崎、松元}}

As experimental systems, we have constructed a two-projector system and
single-projector system. 
%In the proposed 3D measurement system, two projectors are used to retrieve the 
%ratio of a {\it light flow} at each pixel to estimate depth from a camera. 
%The projectors and 
For two-projector system, 
the projectors and the camera are positioned so that the camera can observe overlapping patterns 
projected by the projectors.
The two 
projectors project 
%sinusoidal patterns (削除) or 
uniformly-spaced parallel lines and
because discrimination of the two 
patterns is required for reconstruction,  different colors are used. 
%in our implementation.
 The setup with two 
projectors and a camera is shown in Fig.~\ref{fig:actualequipment}(b). 
For single-projector system, 
one projector projects two non-uniformly-spaced parallel line sets with 
different color. The setup with single 
projector and a camera is shown in Fig.~\ref{fig:actualequipment}(a). 

For both systems, the projected patterns are static and do not 
change; thus, synchronization is not required. 
The camera and 
projector(s) are assumed to be calibrated (\ie, the intrinsic parameters of the 
devices and their relative positions and orientations are known). 
Depths are estimated by the light flow analysis as already described. 
%and the 
%system estimates the depth of each point of the surface of the target object by 
%retrieving the {\it light flows} of two projected patterns at the giving point. 

\subsection{Light flow estimation 
%using parallel lines 
with two projectors}
\label{sec:pararellinetolightflow}
In the analysis of section \ref{sec:depthfrom2proj}, 
pattern motion on the projectors' image planes 
($\Delta q_{h1}$ and $\Delta q_{h2}$) were used. 
%
%As long as the ratio of {\it light flow} on the projector image plane between 
%two projectors is retrieved, any 
%pattern can be used in theory. 
%
%However, 
%%we cannot  
%%transform the pattern motion on the camera image plane to that on the projector 
%%image plane without a 
%%surface normal for each pixel, 
%%and estimation of the surface normal from one or 
%%two images is difficult. 
%these values are not directly observed by simply tracking the apparent motion
%of the patterns in the captured images. 
To observe these values, 
we propose to use 
\fcut {
two algorithms that
directly calculate the ratio  from the 
observed pattern without performing transformation.
%we need to either estimate surface normal of object 
%or calculate the ratio without conducting transformation.
%Since estimation of surface normal from single or two images is known to be 
%impossible, we propose two solutions to directly retrieve the ratio from the 
%observed pattern. 
The first solution uses a sinusoidal pattern to estimate the 
phase at each position. Because the phase is the same for both the camera and  projector image planes, transformation is not required.
The second solution uses 
}
a uniformly spaced line pattern with constant intervals 
as ``scale'' for measuring the pattern motion.
In this method, the length of blur is retrieved from the observed image as
the width of the band of blurred lines. Then, the ratio of the blur on the projector 
plane can be calculated by dividing the length by the line intervals on the same image.
Since the line intervals are constant on the projector's image plane, 
the ratio 
%of the apparent blur width and the apparent line intervals 
%is 
becomes
a pattern motion on the projector's image plane.

The blur of the parallel lines are observed only for each lines,
thus, the resolution of this approach is relatively sparse. 
However, it has a great potential to reconstruct extremely fast motion
using only a  single image.

The method is as follows:
%We can estimate {\it light flows} $\Delta q_{h1}$ and $\Delta q_{h2}$ by
%projecting a pattern of vertical lines in uniform intervals.
The projected pattern is a set of parallel vertical lines. 
The captured lines projected onto the object
move with the motion of the object.
The exposure time of the camera is assumed to be set
so that the apparent motion of the vertical line of the image 
is observed as motion blurs of the lines.

Let the width of the motion blur of the line defined as  $\Delta r$.
$\Delta r$
cannot be used directly as 
$\Delta q_{h}$ in Eqn.~\ref{eqn:deffarential},
because
$\Delta q_{h}$ is a displacement on the projected pattern image,
whereas
$\Delta r$
is a displacement on the captured camera image.
%
%However, 
If vertical lines at uniform intervals are projected,
$\Delta r$
can be normalized by the apparent intervals on the camera image
and 
can be used as 
$\Delta q_{h}$. 
If the apparent motion blur of the line 
on a local patch in the camera image 
is
$\Delta r$
and the 
interval between the lines on the same patch 
is 
$B$,
then $\Delta q_{h}$ can be approximated as follows:
\begin{equation}
%\Delta q_{h} \approx  \frac{\Delta r}{B}.
\Delta q_{h} \approx  {\Delta r}/{B}.
\label{eqn:flowestimateline}
\end{equation}

\fnoteIII{
In the current implementation, 
the blur is detected just by simple adaptive binarization. 
Then, $\Delta r$ and $B$ of Eqn.~\ref{eqn:flowestimateline} are estimated with
sub-pixel accuracy by localizing crossing positions of the profile
with the threshold levels of the adaptive binarization 
(the positions marked by red circles in Fig.~\ref{fig:flowestimate-line}(e)).
}

By projecting vertical lines from two projectors
in different colors (\eg, red and blue)
and by estimating 
$\Delta q_{h1}$ and $\Delta q_{h2}$ from the different color channels,
depth estimation from motion blur becomes possible. 
Fig.~\ref{fig:flowestimate-line} shows the pattern images in (a, b),
projected line patterns with motion blurring in (c),
an apparent line interval (\ie, $B$) and a size of blur
(\ie, $\Delta r$) in (d), 
and the {\it light flow} estimation 
by interpolating ${\Delta r}/{B}$.
%using equation
%(\ref{eqn:flowestimateline})
%and for $x$ axis.

\fnoteIII{
If the target surfaces have object boundaries, the line patterns cannot be detected
outside the boundaries. 
Thus, blurred patterns are disconnected
and values of ${\Delta r}$ or ${B}$ become abnormal at those points. 
Since the assumption of smooth surfaces are not fulfilled there, 
we remove these regions. 
In the current implementation, we specify the upper and lower limits
for the intervals $B$ and label
pixels as outliers where $B$ exceeds this range.
The outlier points are removed from the 3D reconstruction. 
The boundary points are removed from the reconstruction process
using this technique.  
}

\fnoteIII{
The line directions of the pattern are not required to be perpendicular to
scan lines. This is because although blur widths $\Delta r$
and line intervals $B$ along scan lines are affected by the
apparent line directions, the ratio ${\Delta r}/{B}$ are
not affected by the directions. 
}

In this method, 
the spatial resolution of the {\it light flows} is as coarse as the apparent line intervals, 
but fast-moving objects that are only observable as motion blur can be measured. 

\fnoteIII{
About the precision of the method, 
we can conduct a coarse analysis on depth precision based on 
Fig.~\ref{fig:depthfigratiofunc}(a). 
Let $\Delta r$ and $B$ be 10 and 30 pixels, which is a
typical setup of the experiments shown later. 
Also, let precisions of those
values be $1/4$ subpixels. 
Then, the errors of $\log{(\Delta r)/B}$ 
is approximately 0.049
(we take $\log$ because $h^{-1}$ takes $log$ values). 
In Fig.~ref{fig:depthfigratiofunc}(a), 
this is about 0.02-0.03m depth errors at 0.5m distance, and about 0.06m
depth errors at 1.0m distance.
}

%等間隔のvertical linesを投影し、運動する物体を撮影する場合を考える。
%カメラから撮影される、投影されたveritical linesは、物体の運動によって移動する。
%veritical linesの見かけの移動速度が、シャッター速度のexposure timeによって、
%blurとして観測可能であるとする。
%この時の、
%カメラ画像上での見かけのblur幅を
%$\Delta r$
%とする。
%$\Delta r$
%は、
%式\ref{eqn:deffarential}
%の
%$\Delta q_{h}$
%として直接は利用できない。
%なぜなら、
%$\Delta q_{h}$
%パターン画像上での変化量であり、
%$\Delta r$
%はカメラ画像上での変化量であるからである。

%ただし、
%等間隔のvertical linesを投影する場合、
%カメラ画像上で観測されるvertical lineのintervalによって
%$\Delta r$を正規化し、
%$\Delta q_{h}$
%として利用することが出来る。
%画像上のある点の周囲において、
%見かけ上のblur幅が
%$\Delta r$であり、
%平行線の間隔が
%$B$であるならば、
%\begin{equation}
%\Delta q_{h} \approx  \Delta r / B
%\end{equation}
%の近似が可能である。

%２個のプロジェクタから、赤、青の２種類の色でvertical linesを投影し、
%それぞれで観測されたblur幅$\Delta r_1$,$\Delta r_2$,
%平行線幅$B_1$,$B_2$から
%$\Delta q_{h1}$,$\Delta q_{h2}$を計算することで、
%depthを計算できる。
%この方法では、
%画像上でのdepthの解像度は、プロジェクタの線の幅程度になるが、
%blurしか観測できないような速い動きの物体の奥行きを推定することが出来る。

\subsection{Light flow estimation 
%using parallel lines 
with a single projector}

%\subsection{Special case of one projector and two patterns}
%これまでに、２個のパターンを２個のプロジェクタで投影することで、
%depth estimationを行う方法について解説した。
%説明した方法における特殊な場合として、
%二つのパターンを単一のプロジェクタから投影した場合を考える。
%このとき、degenerate conditionが生じて、depth推定が不可能になる。
%具体的には、$h(d)$がconstant functionになり、
%$h^{-1}$が定義できない。

We have explained depth estimation method with two patterns projected
from two projectors. 
Now, we consider the special case of two patterns, where the two patterns
are projected from a single projector. 
In this case, depth estimation becomes impossible because of degenerate conditions.
Specifically, the function $h(z)$ becomes constant functions for all  the camera pixels, 
thus, $h^{-1}$ cannot be defined.

%このような場合でも、二つのパターンの間隔に、異なるmodulationをかけることで、
%depth estimationが可能になる。例えば、
%パターンの間隔が、プロジェクタの左端と右端で、2倍異なる間隔になるように
%パターン(Fig\ref{fig:nonuniformpat})を作成する。
%このようなパターンは、画像のhorizontal coordinate $p$が$[0,1]$の範囲にある場合、
%$p_0^{new} = 2^{p_0} - 1$ によって、$p_0$を$p_0^new$にmappingすれば、
%作成できる。
%逆に、もう一つのパターンについては、これと左右逆のmodulation
%$p_1^{new}=2-2^{1-p_1}$をかける。
%このようなsetupの下で、$h(z)$の計算時に、
%画像座標に、逆のmapping、つまり、
%$p_0 = \log _2 (p_0^{new} + 1)$,
%$p_1 = 1-\log _2 (2-p_1^{new}) $,
%をかけることで、constantでない$h(z)$を得ることができる。
%このtechniqueにより、同じアルゴリズムで、単一プロジェクタによる
%depth推定を実現できる。

\begin{figure}[tb]
\vspace{-5mm}
\begin{center}
	\includegraphics[width=0.16\textwidth]{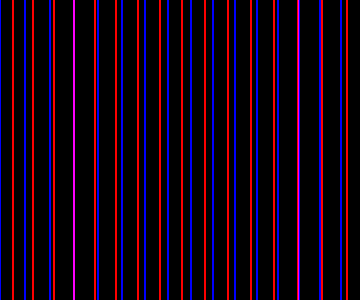}
	\includegraphics[width=0.16\textwidth]{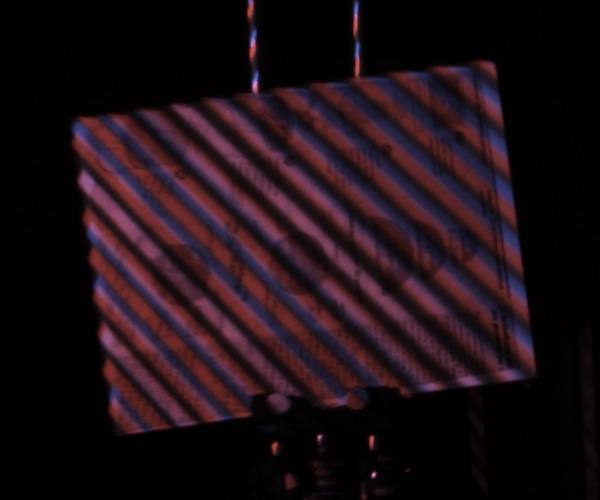}
	\caption{Left: Modulated nonuniform parallel line patterns and Right: captured image 
    with strong motion blur.}
	\label{fig:nonuniformpat}
\vspace{-5mm}
\end{center}
\end{figure}

Even in this case, by applying different modulation for these patterns, 
depth estimation becomes possible. 
For example, we modulate the pattern intervals so that the 
intervals becomes wide by two times at the right side with respect to the left side 
(Fig.~\ref{fig:nonuniformpat}).
This modulation is achieved with the mapping of the horizontal coordinate $p_0$
by 
$p_0^{new} = 2^{p_0} - 1$,
where range of $p_0$ is $[0,1]$. 
On the contrary, 
for the other pattern, 
intervals are modulated by the mapping reversed for the left and right side:
$p_1^{new}=2-2^{1-p_1}$.
With this setup, 
in calculation of $h(z)$, 
the inverse mappings
$p_0 = \log _2 (p_0^{new} + 1)$,
$p_1 = 1-\log _2 (2-p_1^{new}) $,
are applied.
Then, non-constant $h(z)$ can be obtained. 
With this technique, 
we can estimate depth using two patterns projected from a single projector
with the same algorithm as the two-projectors case.

\begin{figure}[tb]
\vspace{-5mm}
\begin{center}
	\includegraphics[width=0.21\textwidth]{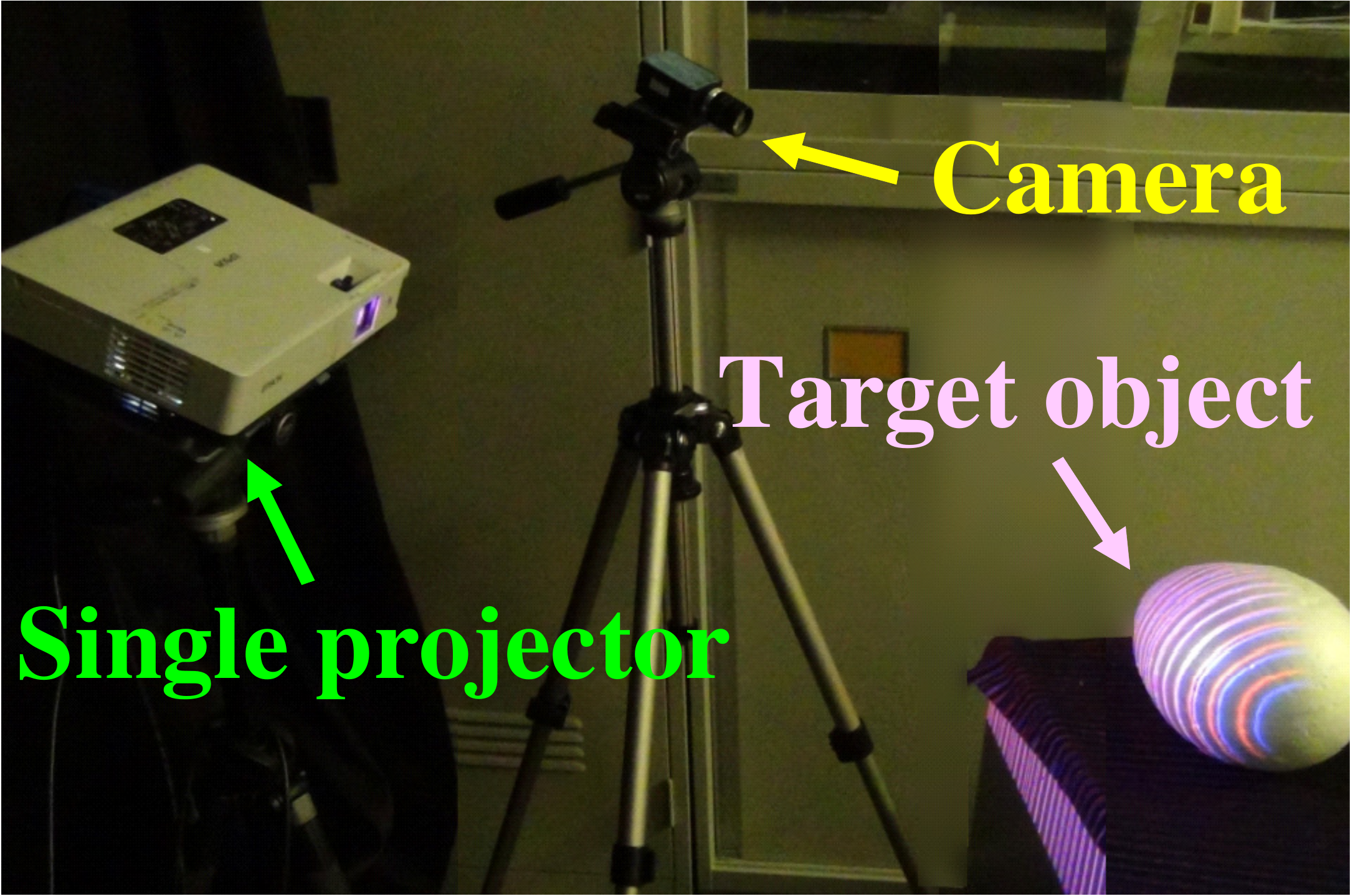}
	\includegraphics[width=0.21\textwidth]{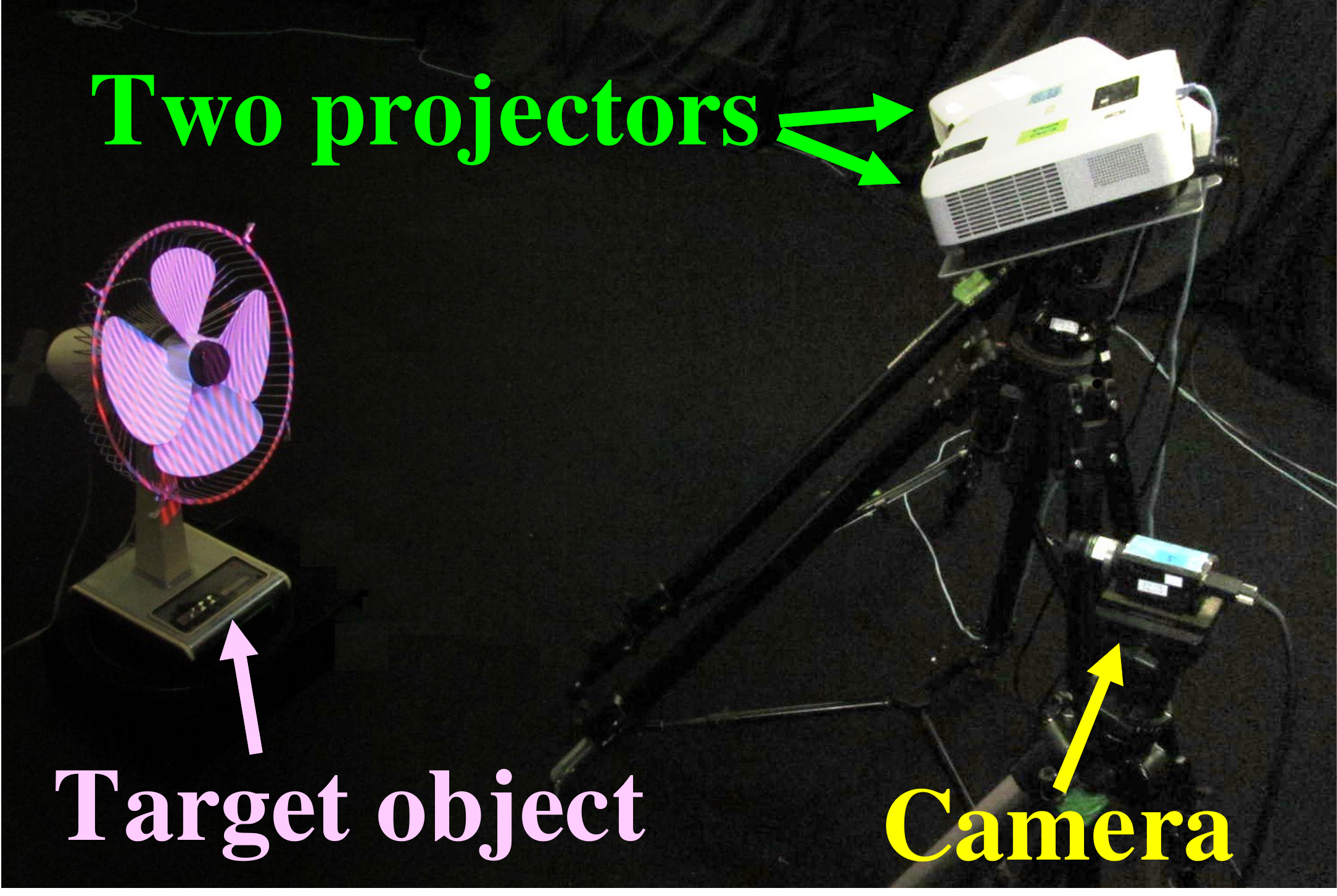}\\
(a)\hspace{40mm}(b)
	\caption{Two examples of the system configuration. (a) single projector 
    projects non-uniformly spaced lines with two colors and (b) two projectors 
    project uniformly spaced lines to make overlapping pattern on the object.}
	\label{fig:actualequipment}
 	\vspace{-5mm}
\end{center}
\end{figure}

\begin{figure}[tb]
%\vspace{-5mm}
\begin{center}
	\includegraphics[width=0.45\textwidth]{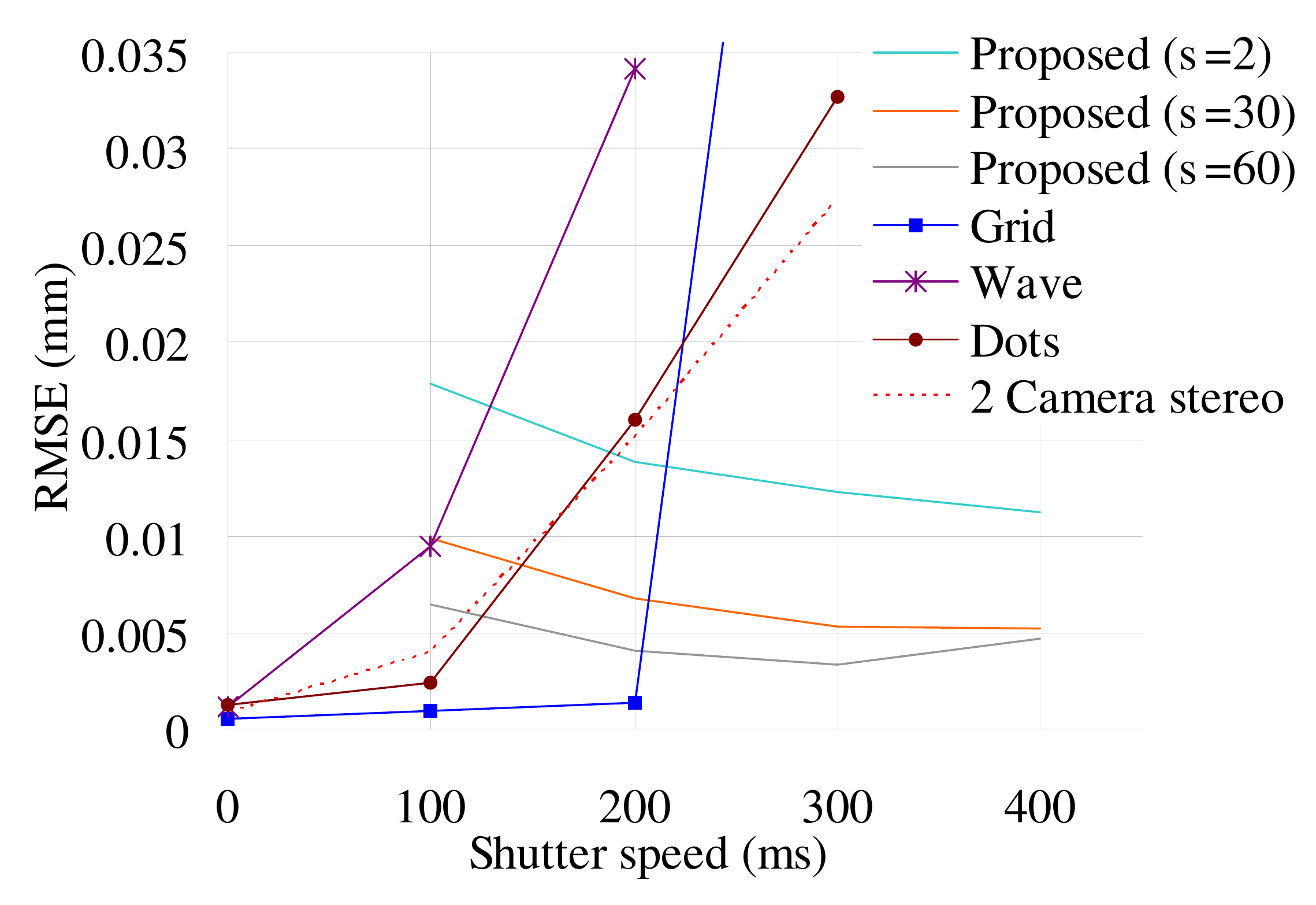}\\
	\caption{Accuracy evaluation for captured plane with line  patterns. 
	Vertical axis are RMSEs and horizontal axis are shutter speeds. 
	Since {\it light flows} are approximated by piecewise linear interpolation as Fig.~\ref{fig:flowestimate-line}(e),
	they are smoothed with Gaussian filters with $\sigma=2, 30, 60$.
	The size of the board is $600 \times 400 mm$.}
	\label{fig:Graph_line}
	\vspace{-5mm}
\end{center}
\end{figure}

\section{Experiments} %\knote{川崎、松元}}

\subsection{Evaluation with planar board}

The first experiment was conducted 
with two-projector system
(Fig.~\ref{fig:actualequipment}(b)).
%by using two projectors and a camera system shown in 
%Fig.~\ref{fig:actualequipment}(b). %{fig:HalfMillar}(b).
%Images were captured by shifting the target screen placed on a motorized stage.  
%Because of the limitation of the length of the motorized stage, we put a close-up 
%lens to change the scale as to be 1/3 of real length. 
%In the set-up, 
%the 
%motion range of the screen was 500mm-800mm from the projector and the camera, 
%in-focus distance was 650mm $\pm100$mm for the projector, and the reference plane capturing interval was 1.0mm.
The camera resolution was $1600 \times 1200$ pixels, projector resolution was 
$1280 \times 800$ pixels, and the baselines between the 
camera and the two projectors were both approximately 400mm.
%; note that no 
%baseline is essentially required between projectors with our technique, which is 
%equivalent to the single projector set-up.

For evaluation, we attached a target board onto the motorized stage, and captured it with
the camera while the board was moving under
different conditions, such as different shutter speed and different 
projection pattern.
%, and
%different material of the target object.
%Under the condition, a total of 14 images ware captured.
The RMSE was calculated after plane fitting to the estimated depth.
Results are shown in 
%Figs.~\ref{fig:Graph_phase}  and 
Fig.\ref{fig:Graph_line}.

%Estimated depth was calculated by averaging 1000 points of estimated depths of 
%each frame. 
%In the graph, we can observe only our method and 
%Graycode~\cite{Inokuchi:ICPR84} can recover the correct depth for all the ranges. 
%Further, Graycode method gradually reduces its 
%precision at out-of-focus zone, whereas our method keeps almost same accuracy for all the ranges.

%In the first graph of Fig.~\ref{fig:Graph_phase}, we can see that an
%increase in shutter speed does not affect the RMSE. In addition, different 
%texture and material does not affect the results; we use newspaper and cover 
%of magazine for experiment.
%As shown in Fig.~\ref{fig:Image_phase}, we can confirm that the longer shutter speed
%makes larger blur on sinusoidal pattern, however, phases are correctly estimated 
%with our phase estimation algorithm, which 
%results in maintaining the similar RMSE at all shutter speeds.

%The cases of the regular line pattern are shown 
%Unlike the case of sinusoidal pattern
%where the flow estimation is processed pixel-wise, 
Since the spatial accuracy of the flow estimation with the line pattern is
an approximation and is affected by the apparent sizes of the blur
bands and the apparent line intervals,
%Since the apparent line intervals were 40 pixels, %about 30 pixels and 50 pixels, 
we smoothed the flow estimation with 
Gaussian kernels with $\sigma$ values of 2, 30, and 60 (the apparent line 
intervals were around 40 pixels in the experiment.)
In the results, we can clearly observe 
that the increase in the shutter speed improves the RMSE.
This is because the longer shutter speed makes the longer bands for blur of
lines, which helps to increase the accuracy on estimating the ratio of {\it light flows}.
Although we also captured data with the shutter speed of 500 $ms$, the bands began to overlap
each other and the depth estimation failed.
For the shutter speed of 400 $ms$, 
filtering with $\sigma=60$ did not improve the result
than the case of $\sigma=30$.
This can be because
the spatial resolution in the case of 400 $ms$ shutter speed
was worse than the case of 300 $ms$.

Fig.~\ref{fig:Image_line} shows examples of the actual captured image and 
reconstructed depth maps with our method. As can be seen, 
%although accuracy is not comparable to state-of-the-art techniques, 
depth are correctly reconstructed just from relative information of flows
without coded positions.
We also applied the line based technique to 
textured materials, such as newspaper % and cover of magazine, and RMSE was about 30mm and 95mm, respectively,
and RMSE was about 30mm
for the shutter speed of 300 $ms$.
This is because our image processing is based on simple thresholding and easily 
affected by textures; solution is our important future work.

We also recover the moving planar board with four other methods for comparison, such as grid based reconstruction~\cite{Taubin:3DIM09}, wave based reconstruction~\cite{Sagawa:ICCV2011}, random dot 
based reconstruction, which is equivalent to Kinect~\cite{Kinect}, and stereo-based 
reconstruction using two cameras.
As clearly shown in the graph, longer shutter speed (equivalent to fast motion) drastically decreases the 
accuracy, and no technique can recover meaningful shapes when shutter speed exceeds 300 $ms$.

\fnoteIII{
The above performance decrease of the conventional methods 
is caused by motion blurred patterns. 
}
Fig.~\ref{fig:captured_images} show examples of the projected patterns, the 
real captured images, and the line detection results at the shutter speed of 300 $ms$.
As can be seen, the line detection results become unstable with severely blurred pattern, 
resulting in failure reconstruction.
As for the two camera stereo case, since light flow is view-dependent, which is 
apparently shown in (i) and (j), it fails to find correspondences 
for reconstruction (the sizes and directions of the blur are different).

\subsection{Arbitrary shape reconstruction}

\begin{figure*}[t]%[htb]
  \begin{center}
	\includegraphics[width=0.13\textwidth]{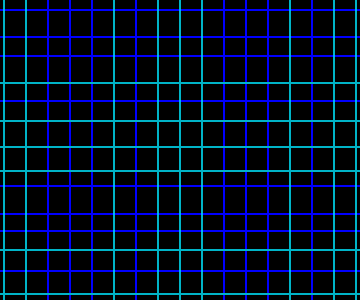}
	\includegraphics[width=0.13\textwidth]{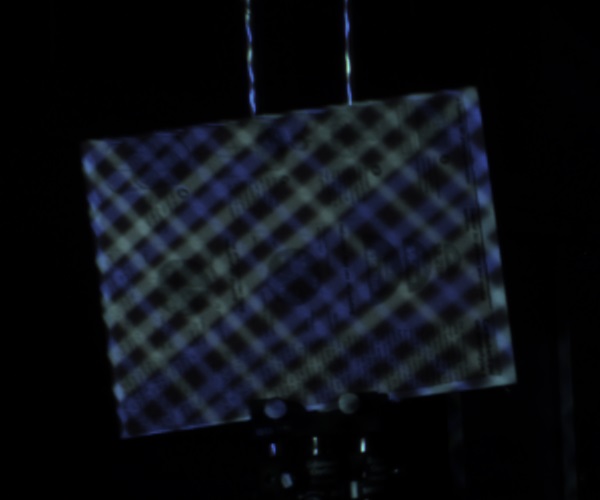}
	\includegraphics[width=0.13\textwidth]{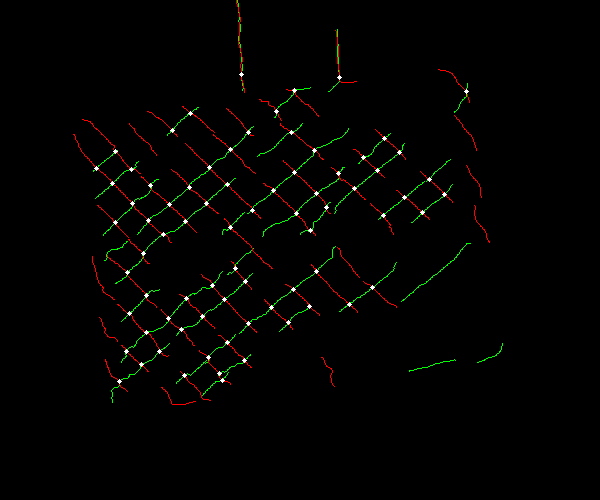}
	\includegraphics[width=0.13\textwidth]{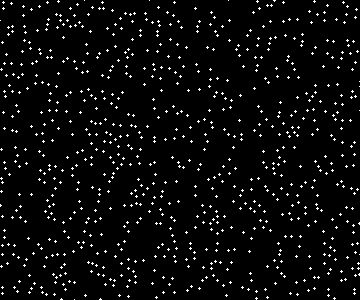}
	\includegraphics[width=0.13\textwidth]{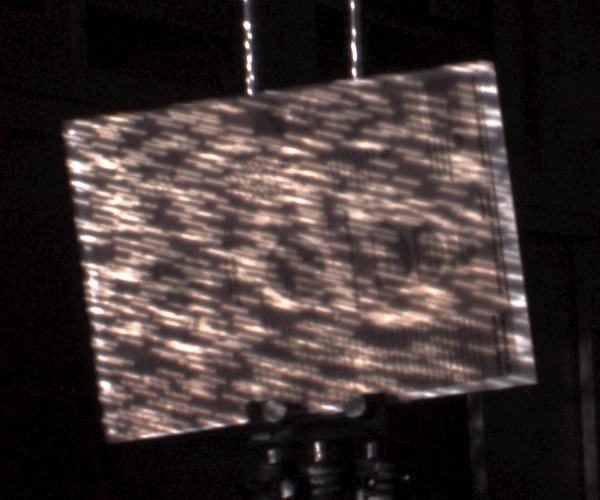}\\
	(a)\hspace{20mm}(b)\hspace{20mm}(c)\hspace{20mm}(d)\hspace{20mm}(e)\\
	\includegraphics[width=0.13\textwidth]{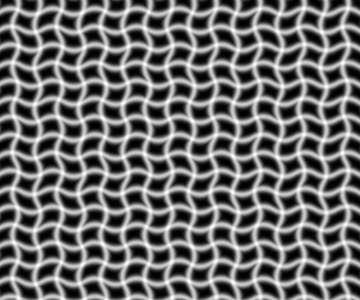}
	\includegraphics[width=0.13\textwidth]{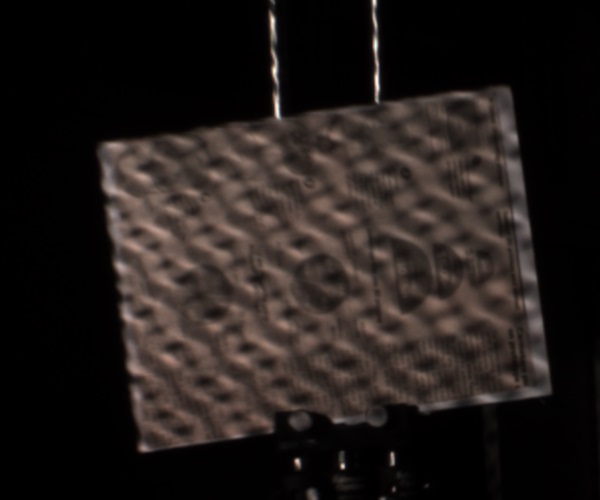}
	\includegraphics[width=0.13\textwidth]{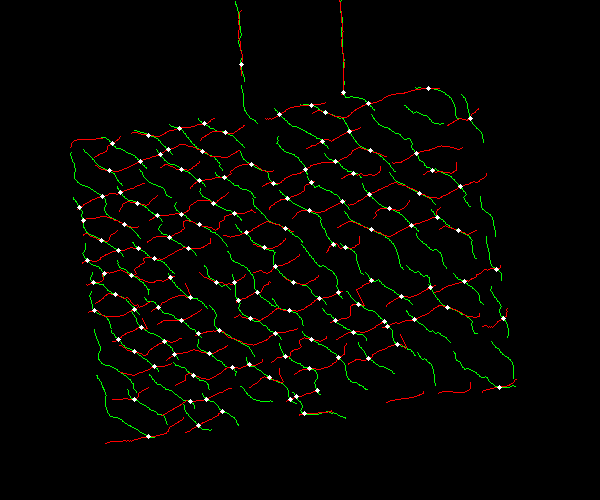}
	\includegraphics[width=0.13\textwidth]{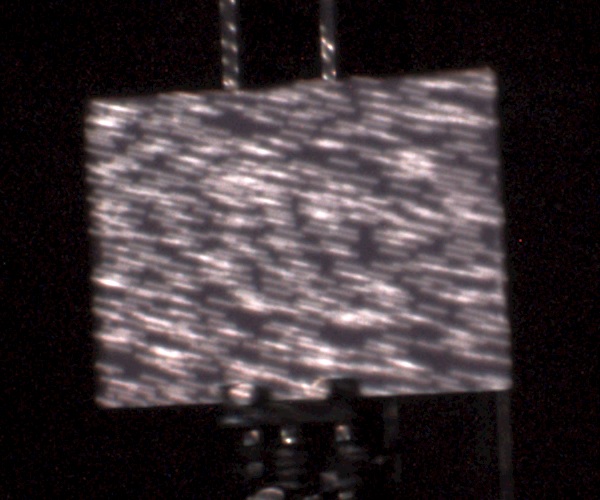}
	\includegraphics[width=0.13\textwidth]{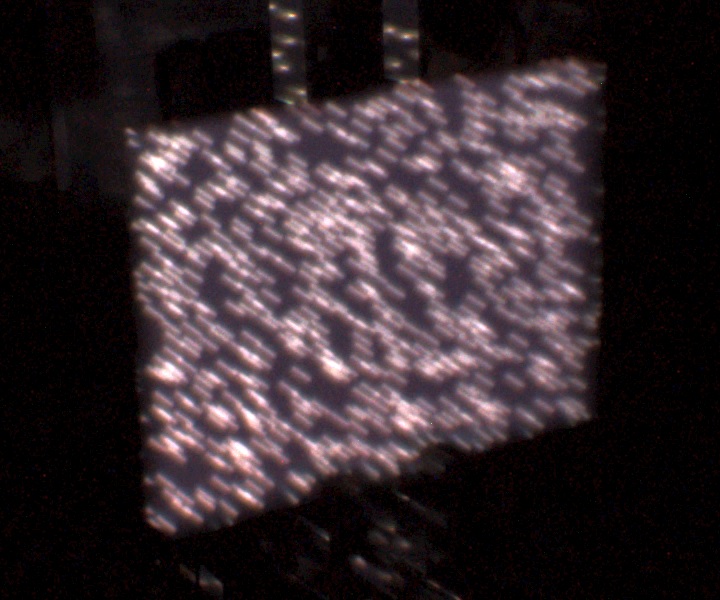}\\
	(f)\hspace{20mm}(g)\hspace{20mm}(h)\hspace{20mm}(i)\hspace{20mm}(j)\\
    \caption{(a) grid pattern, (b) captured image of grid pattern,
% with shutter speed 300 $ms$, 
(c) line detection result of grid pattern, (d) random dot 
      pattern, (e) captured image of random dots, (f) wave pattern, (g) captured 
      image of wave pattern, (h) line detection result of wave pattern, (i) 
      captured image of random dots from the first camera and (j) from the 
      second camera. All the images are captured with shutter speed 300 $ms$.}
    \label{fig:captured_images}
 \end{center}
%\end{figure*}
%\begin{figure*}[t]%[t]
\begin{center}
	\includegraphics[height=26mm]{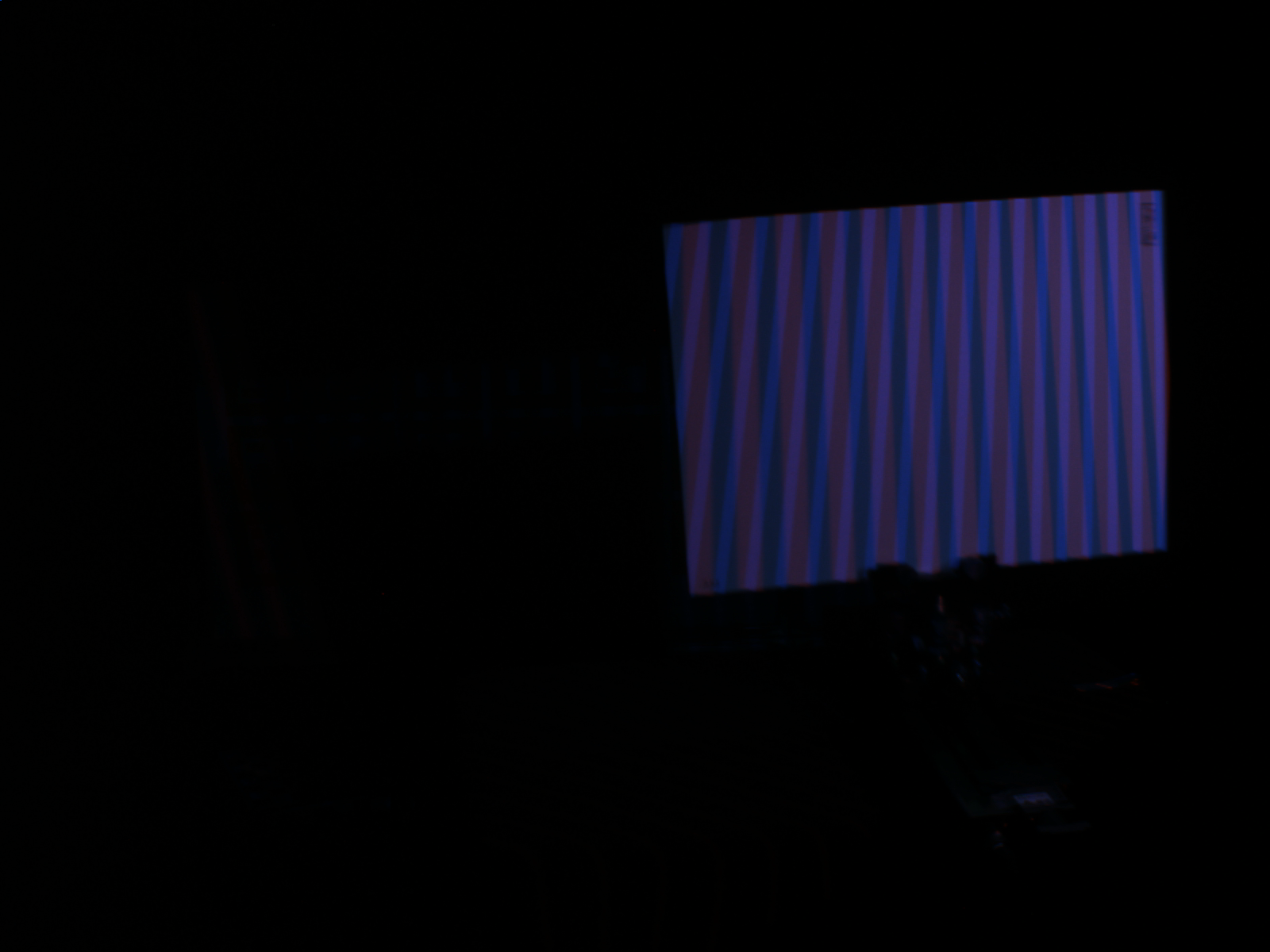}
	\includegraphics[height=26mm]{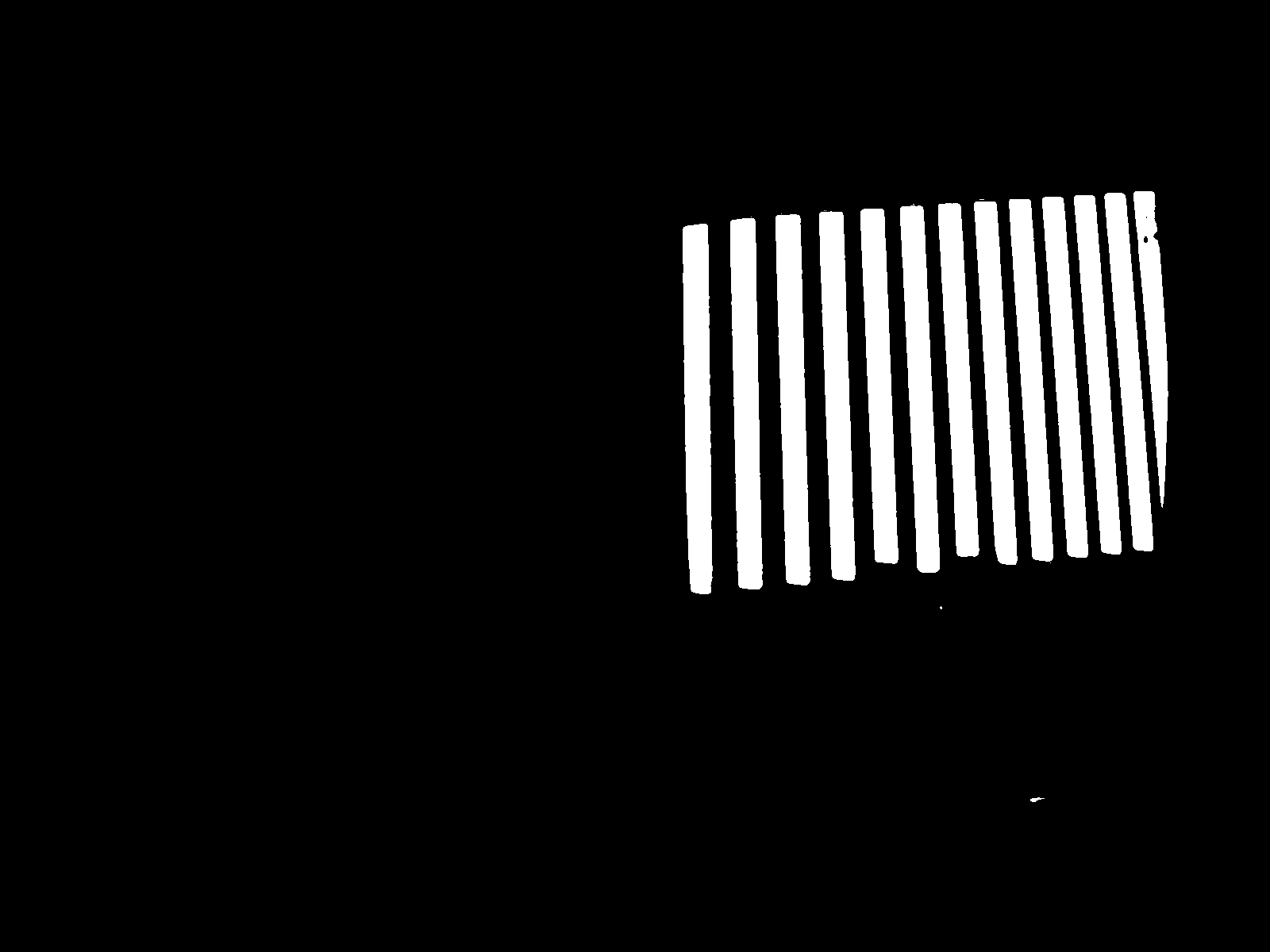}
	\includegraphics[height=26mm]{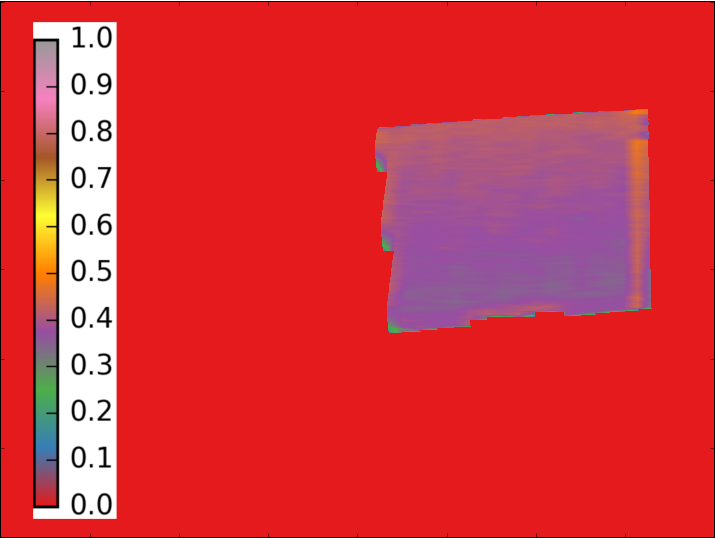}
	\includegraphics[height=26mm]{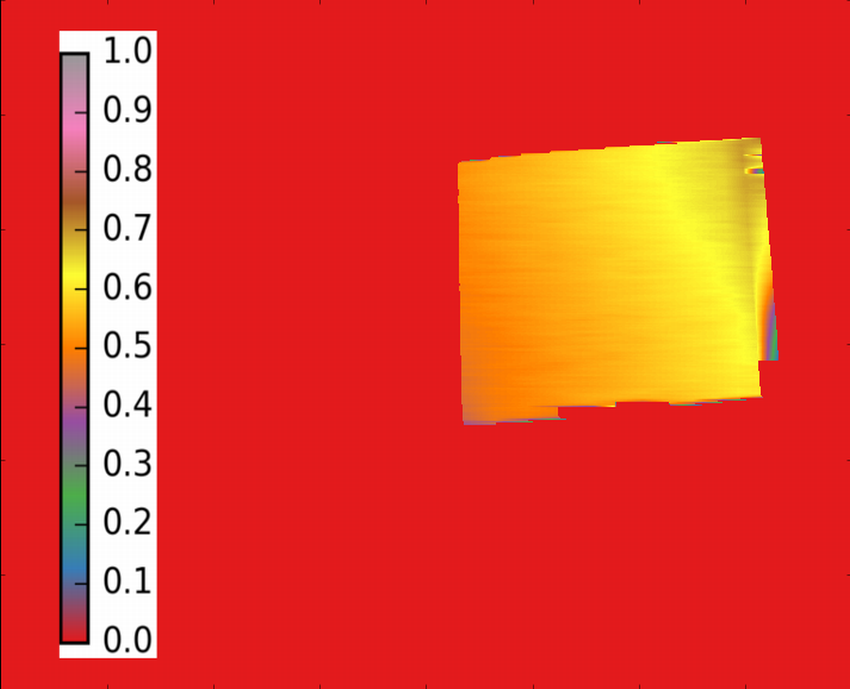}
	\includegraphics[height=26mm]{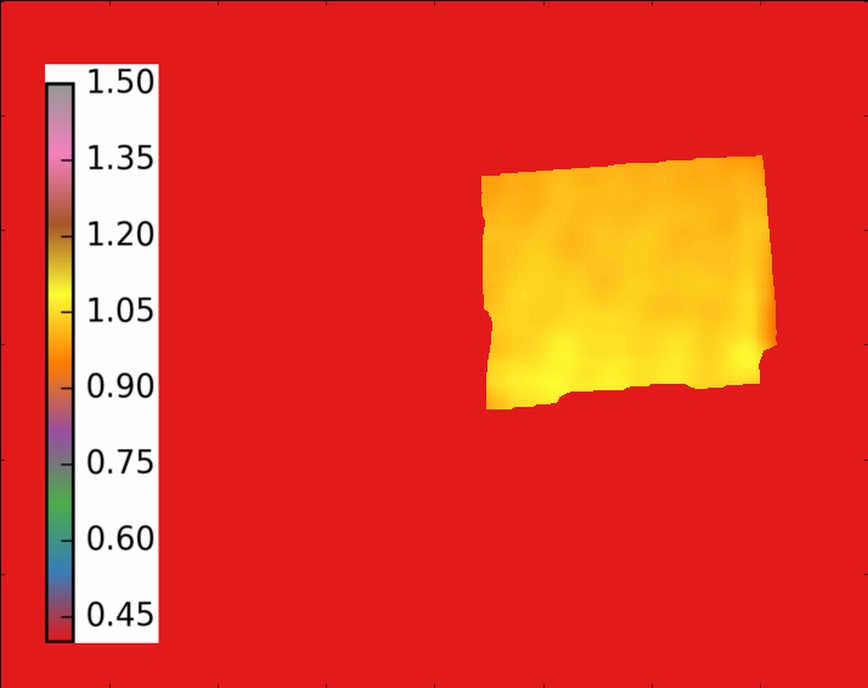}\\
	(a)\hspace{28mm}
	(b)\hspace{28mm}
	(c)\hspace{28mm}
	(d)\hspace{28mm}
	(e)
	
% 	\vspace{10cm}
	\caption{Capturing plane with line pattern for evaluation: (a) captured image, (b) blurred bands (c,d) 
	(c,d) flow estimation of $\Delta q_{h1}$ and $\Delta q_{h2}$, 
%	flow displacements for two projectors,
%	(f) log ratio of displacements (\ie, $h(d)$ of equation \ref{，}),
	and (e) estimated depth.}
	\label{fig:Image_line}
\end{center}
%\end{figure*}
%\begin{figure*}[t]%[p]%[t]
%\vspace{-5mm}
\begin{center}
	\includegraphics[height=27mm]{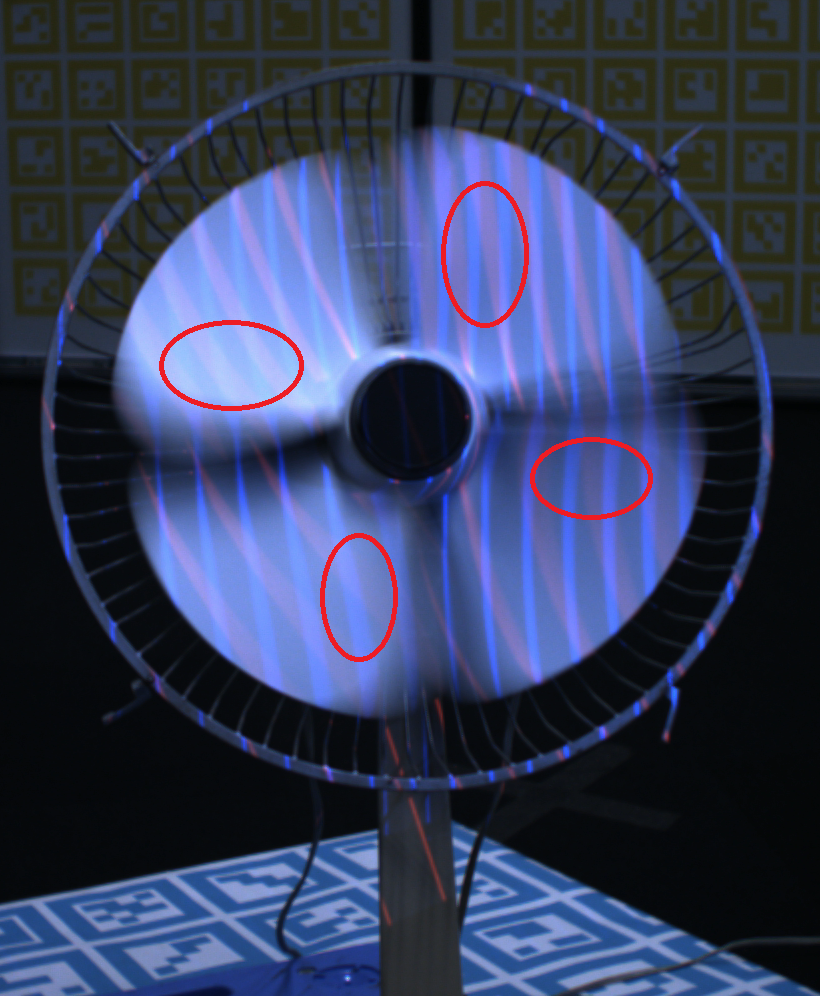}
	\includegraphics[height=27mm]{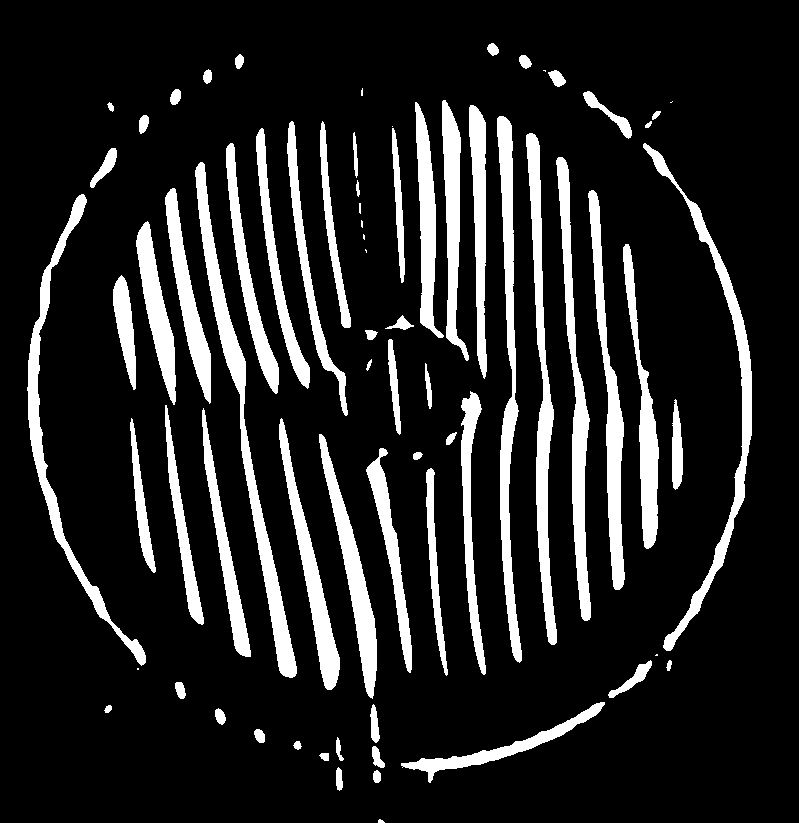}
	\includegraphics[height=27mm]{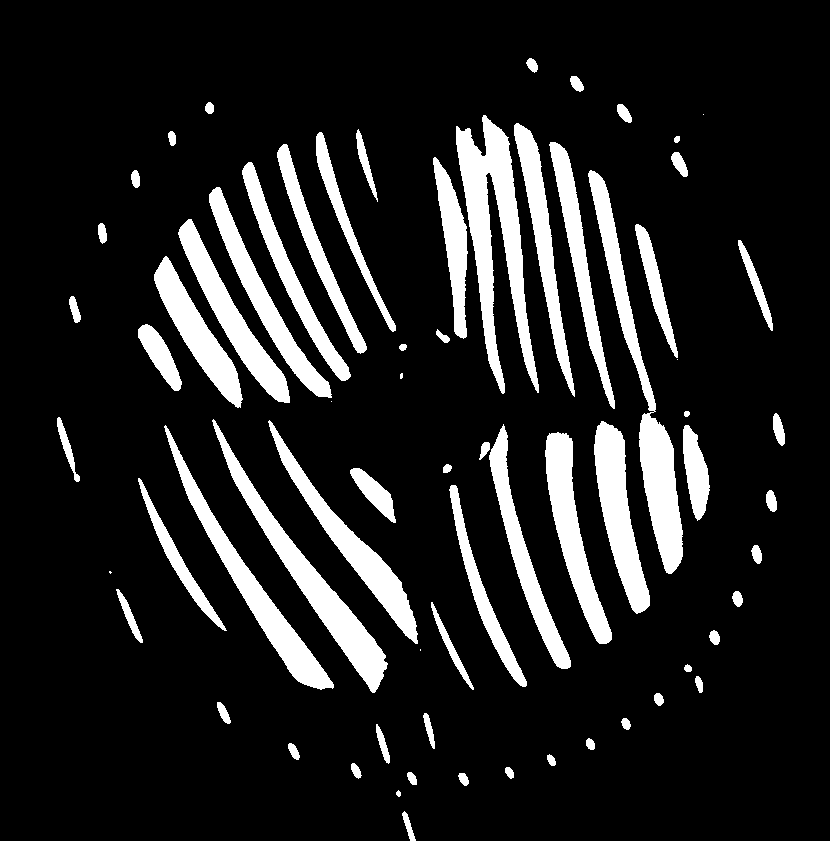}
	\includegraphics[height=27mm]{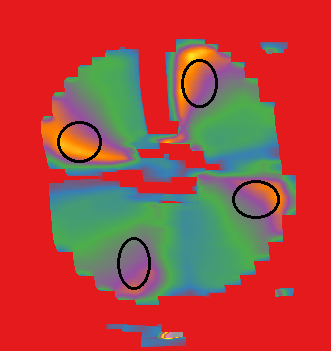}
	\includegraphics[height=27mm]{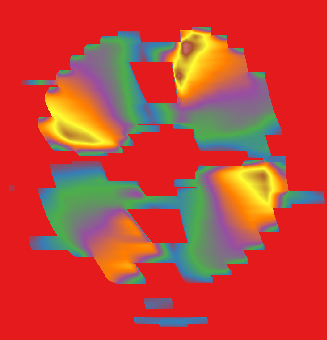}
	\includegraphics[height=27mm]{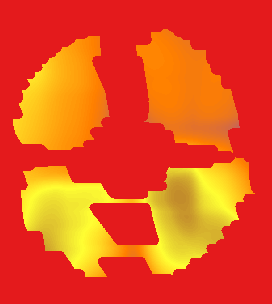}\\
	(a)\hspace{23mm}
	(b)\hspace{23mm}
	(c)\hspace{23mm}
	(d)\hspace{23mm}
	(e)\hspace{23mm}
	(f)
	\caption{Depth estimation of a rotating fan (uniformly spaced line pattern):
	(a) captured image, (b,c) blurred bands (d,e) flow displacements for two projectors,
%	(f) log ratio of displacements (\ie, $h(d)$ of equation \ref{，}),
	and (f) estimated depth. Color mapping of (d,e,f) is the same as Fig.~\ref{fig:Image_line}(c,d,e).}
	\label{fig:capfan}
%\end{center}
%\end{figure*}
%	
%\begin{figure*}[t]
%%\vspace{-5mm}
%\begin{center}
	\includegraphics[height=27mm]{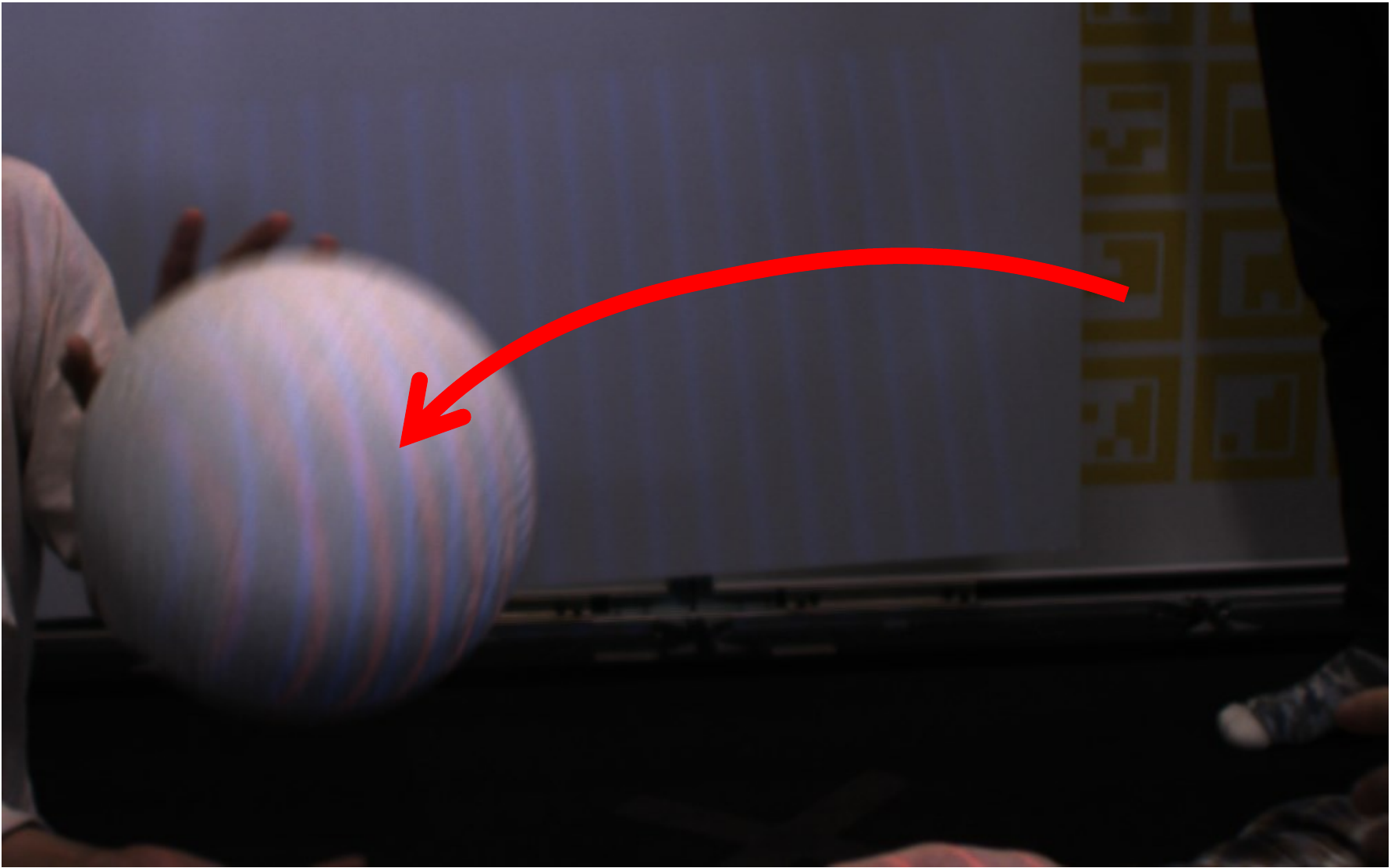}
	\includegraphics[height=27mm]{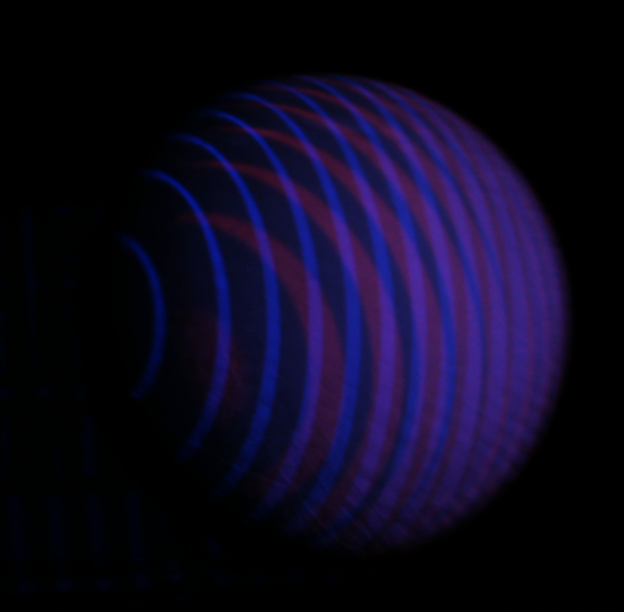}
	\includegraphics[height=27mm]{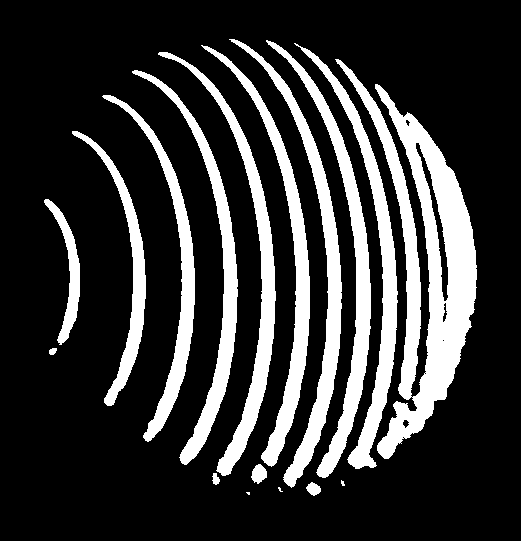}
	\includegraphics[height=27mm]{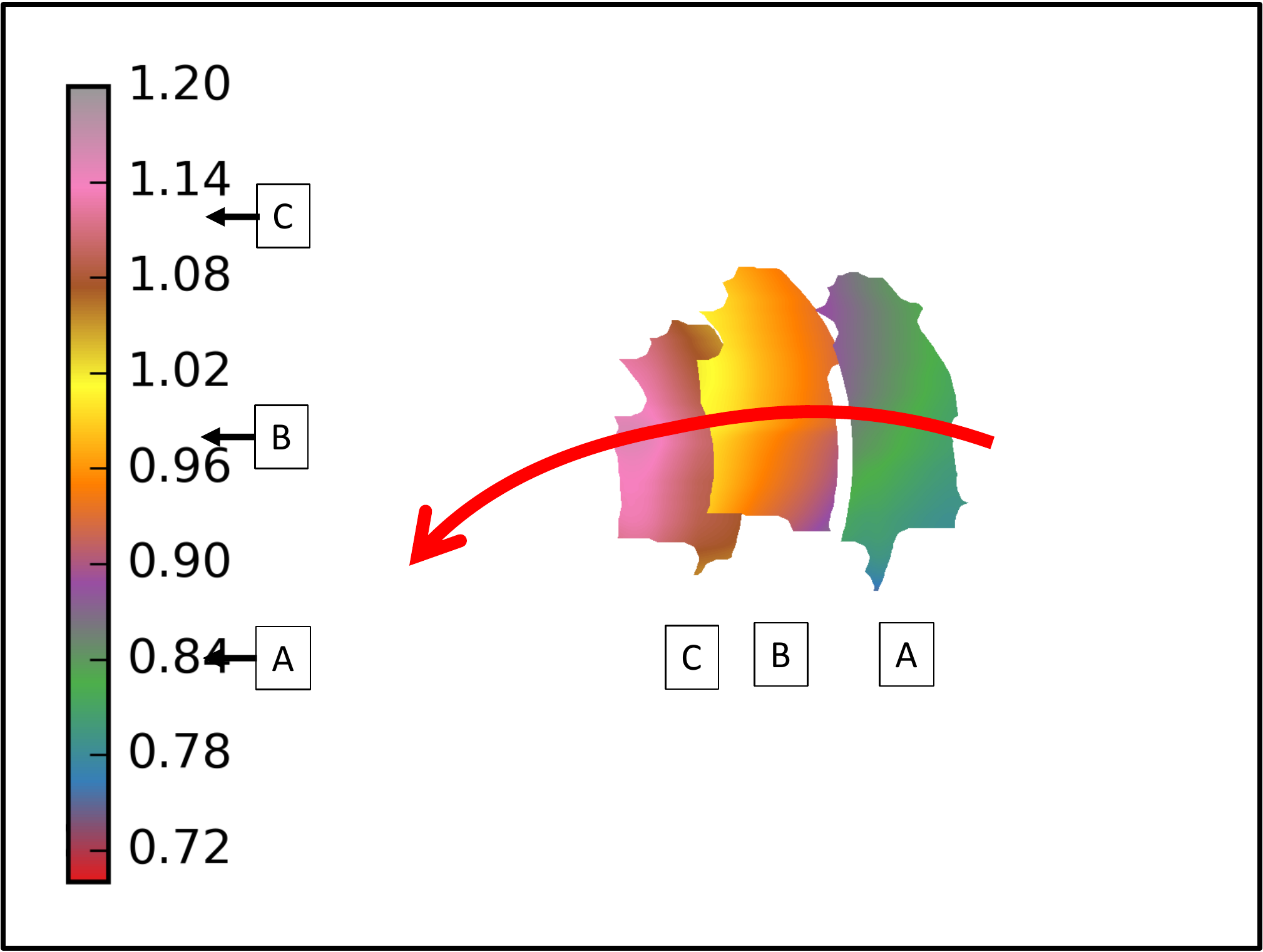}\\
	(a)\hspace{40mm}
	(b)\hspace{20mm}
	(c)\hspace{30mm}
	(d)
	\caption{Depth estimation of a thrown ball (uniformly spaced line pattern):(a) the ball and the trajectory of the throw, (b) the captured image, (c) blurred bands, (d) the estimated depth maps for 3 consequent frames overlapped. Note that, since the image of (a) is blurred, normal structured light methods would fail. }
	\label{fig:capball}
%\end{center}
%\end{figure*}
%
%\begin{figure*}[t]
%%\vspace{-5mm}
%\begin{center}
	\includegraphics[width=0.21\textwidth]{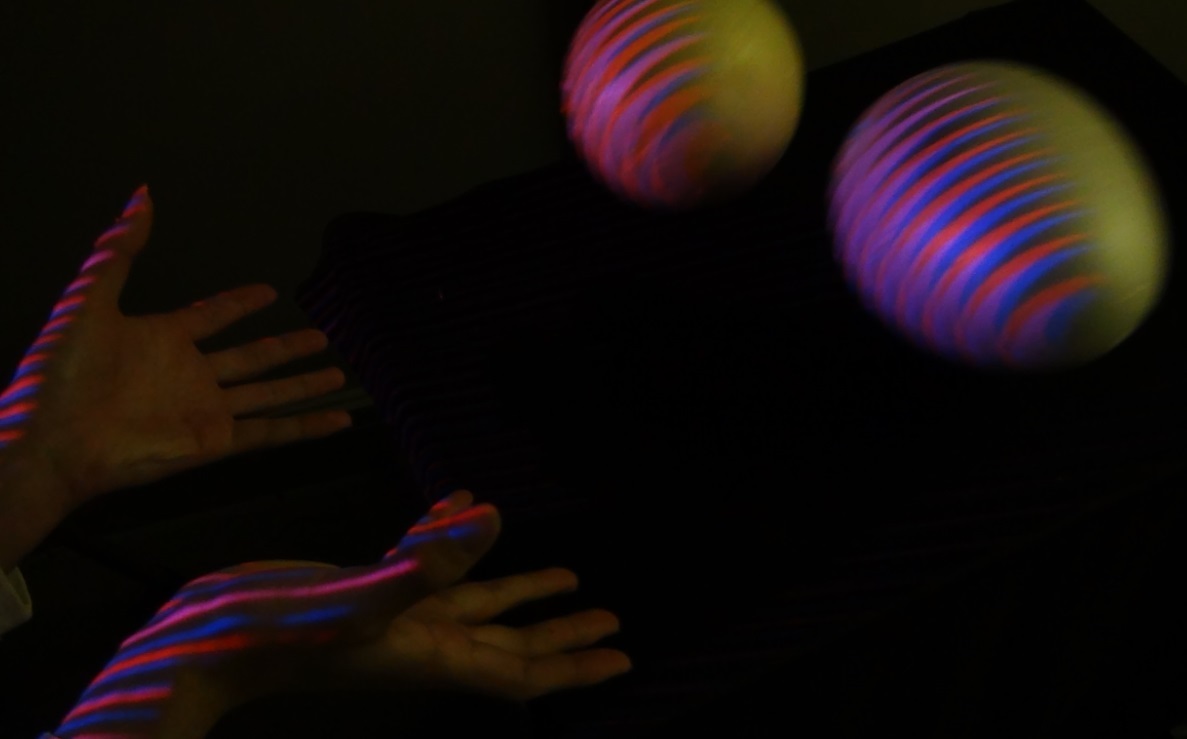}
	\includegraphics[width=0.176\textwidth]{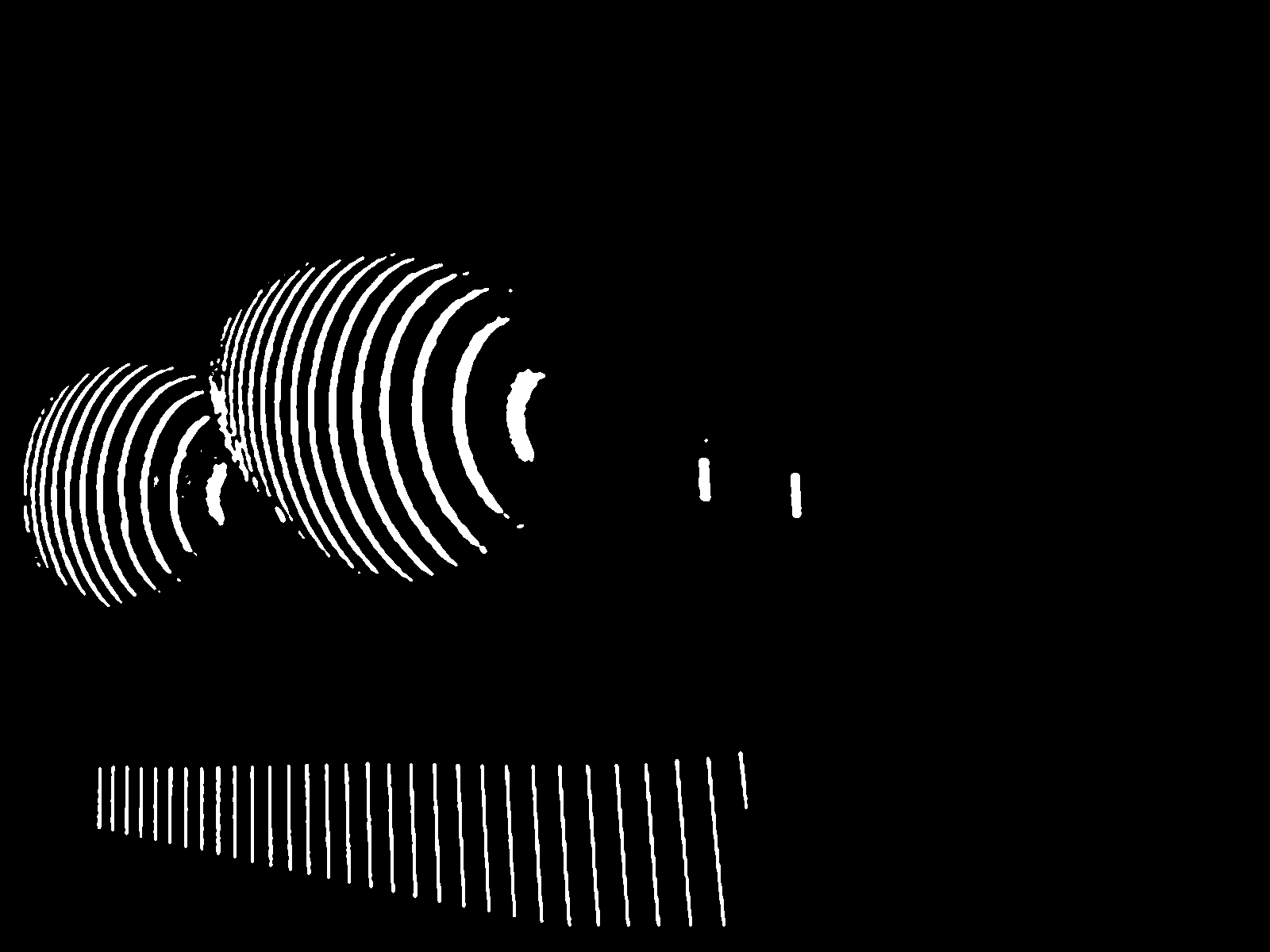}
	\includegraphics[width=0.176\textwidth]{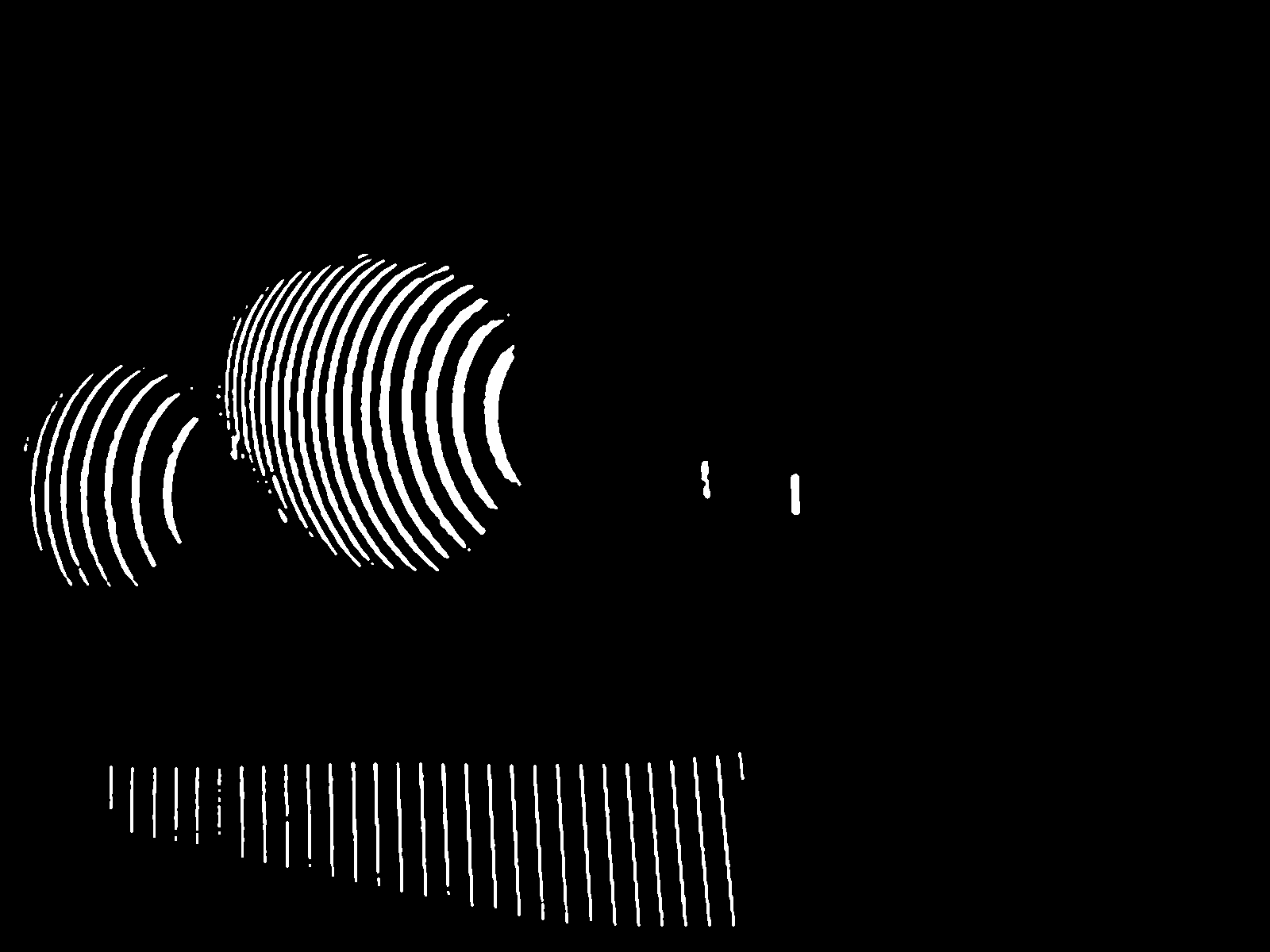}
	\includegraphics[width=0.176\textwidth]{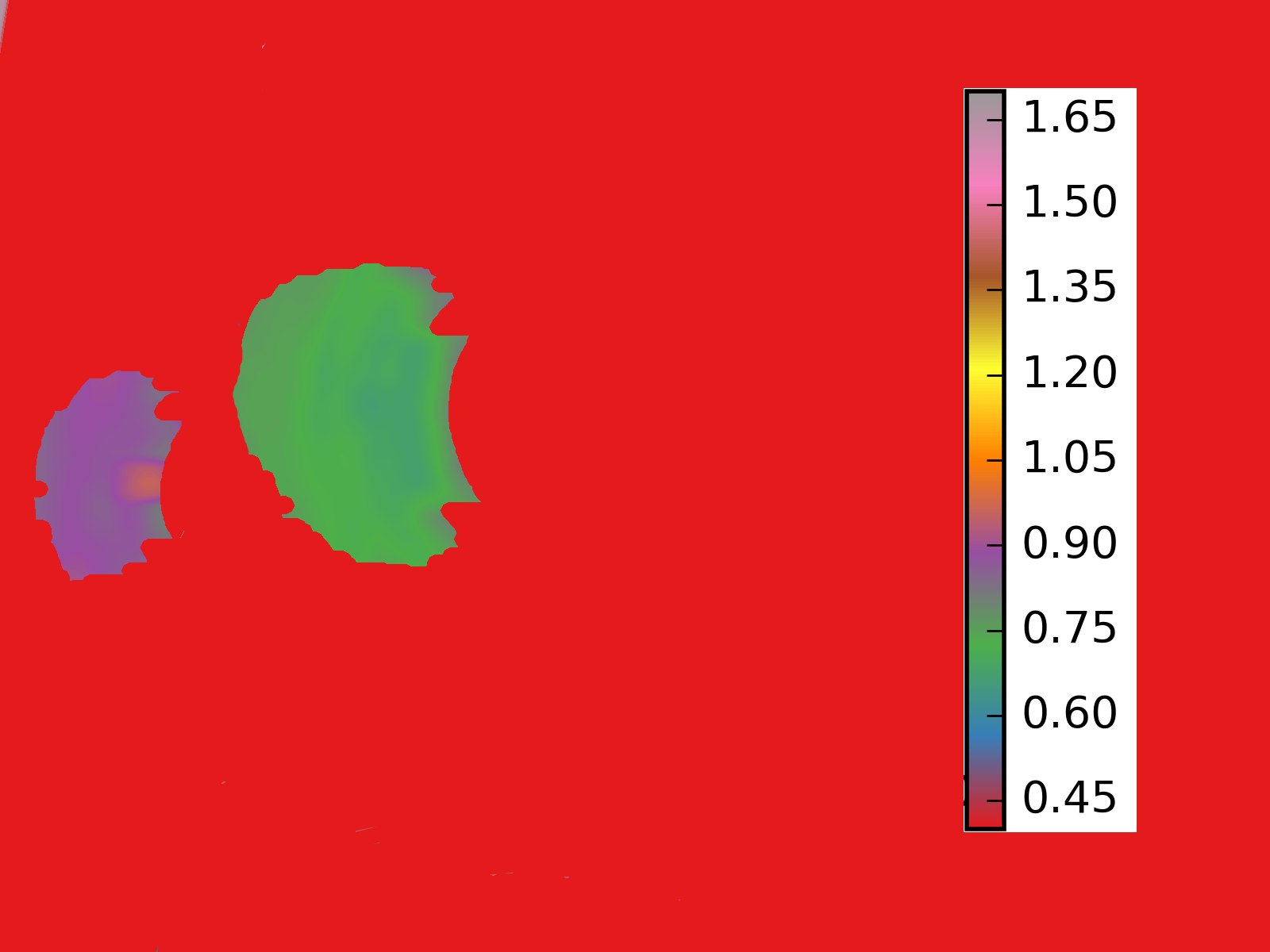}
	\includegraphics[width=0.176\textwidth]{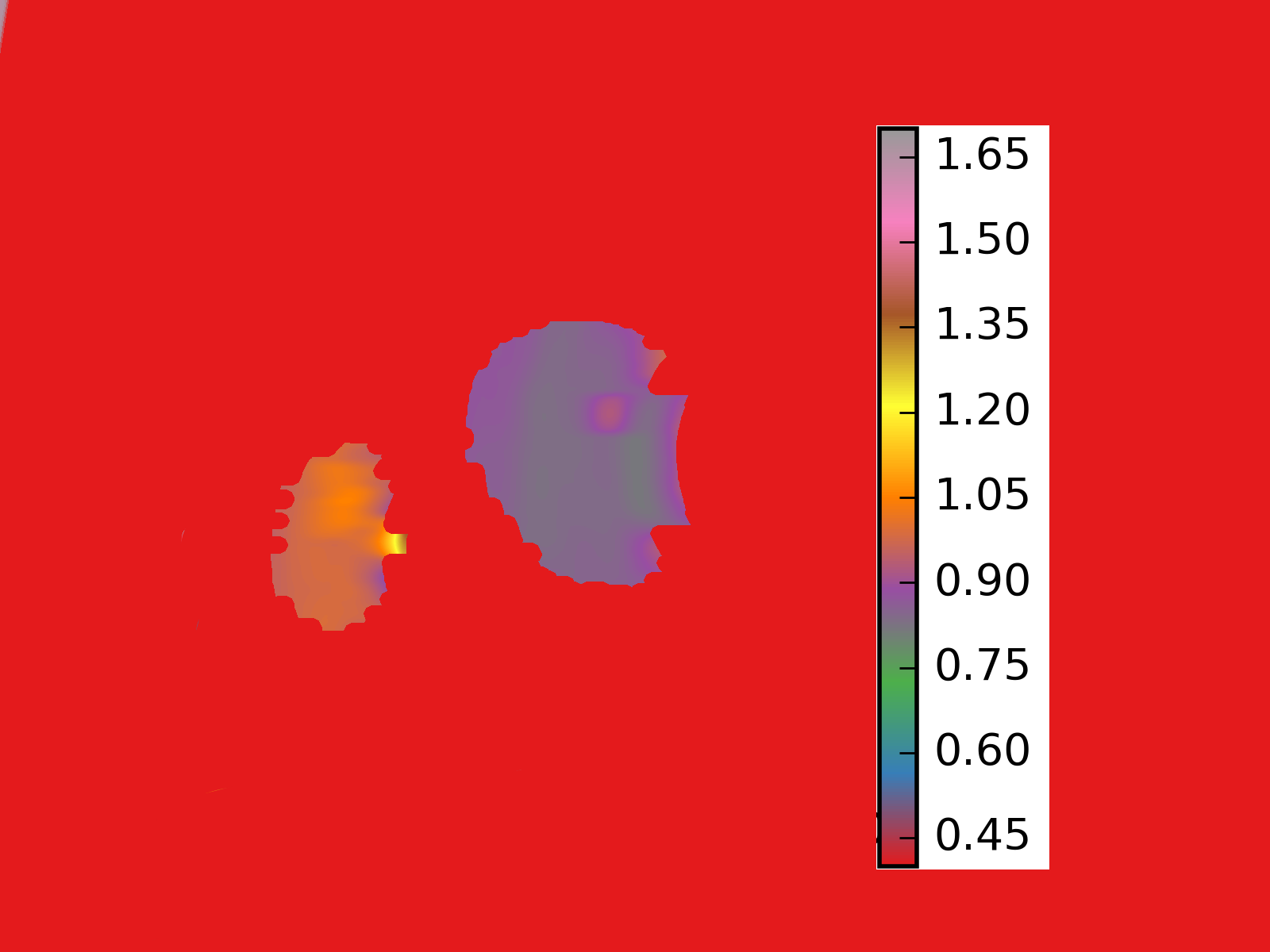}\\
	\hspace{5mm}(a)\hspace{30mm}
	(b)\hspace{25mm}
	(c)\hspace{25mm}
	(d)\hspace{25mm}
	(e)\\
	\caption{Depth estimation of two ball throwing (non-uniformly spaced line 
    pattern):(a) capturing scene of the throwing, (b) extracted red band, (c) 
    extracted blue band, (d) the estimated depth maps for frame 1 and (e) the estimated depth maps for frame 2. }
	\label{fig:twoball}
\end{center}
\end{figure*}

\fcut{

\begin{figure*}[t]
%\vspace{-5mm}
\begin{center}
	\includegraphics[height=27mm]{fig/sinpat-ball-env.png}
\includegraphics[height=27mm]{fig/sinpat-ball-phase0.png}
\includegraphics[height=27mm]{fig/sinpat-ball-flow0.png}
\includegraphics[height=27mm]{fig/sinpat-ball-flow1.png}
\includegraphics[height=27mm]{fig/sinpat-ball-logratio-crop.pdf}
\includegraphics[height=27mm]{fig/sinpat-ball-3d.png}\\
	(a)\hspace{25mm}
	(b)\hspace{25mm}
	(c)\hspace{25mm}
	(d)\hspace{25mm}
	(e)\hspace{25mm}
	(f)
	\caption{Depth estimation of a ball moved by a hand (sinusoidal pattern): (a) the appearance of the ball,
%	 (b) the captured image, 
	 (b) phase (projector 1), 
	 (c,d) {\it light flows} of projector 1 and 2,
	 (e) log ratio of {\it light flows} (\ie, $h(z))$, and (f) 3D points calculated from the distance map for two frames of the image sequence.
	 Color mapping of (c,d) is the same as Fig.~\ref{fig:Image_line}(c,d).}
	\label{fig:caphand}
\end{center}
\end{figure*}

}
We applied our method to object with curved surfaces and fast motion as 
shown in 
%\Fig.\ref{fig:Sample_ShapeRestoration}(left column).
Figs.~\ref{fig:capfan}, \ref{fig:capball}, and \ref{fig:twoball}. %{fig:caphand}.
%Center of the wooden toy is placed 270mm apart from the lens.

The first target was a rotating fan using the uniformly spaced line pattern (Fig.~\ref{fig:capfan}). As can be seen, the blades are blurred even with the fastest 
shutter speed (1ms) 
%was applied 
as shown in (a), whereas a projected pattern made a clear band of 
blur for each line as shown in (b,c).
%the middle column.
Using the detected bands and their width, {\it light flows} are estimated as (d,e),
and depths are correctly reconstructed as shown in (f)
using the calculated ratio of the {\it light flows}. 
Note that, in the flow values shown in (d,e) at the ``front'' parts of the blades (marked by red and black ellipses in (a) and (d)) are larger than the other parts.
Because the fan blades are curved so that the air can be accelerated mostly by the front parts of the blade,
the changing velocity at these parts are the highest.
\fnoteIII{
In this example, 
boundaries of the fan blades are removed from the reconstruction
as outliers as described in Sec.~\ref{sec:pararellinetolightflow}.
}

The second target was a thrown ball using the same setup. Similar to the blades, ball has strong blur on 
captured image (Fig.~\ref{fig:capball}(a)), whereas the projected pattern made a clear band of 
blur for each line ((b,c)).
Using the detected bands, depths are correctly reconstructed,
as shown in overlapped three frames of depth maps in (d).
From (d), the motion of the ball moving from the camera can be clearly observed. 

Finally, two balls were thrown and shapes are reconstructed by a single projector 
setup as shown in Fig.~\ref{fig:actualequipment}(a).
%Since the velocity of the object is not 
%fast, we applied the phase based technique this time.
Fig.~\ref{fig:twoball}(a) shows that projected lines were strongly blurred on the target objects, however two color bands are  
robustly extracted by our algorithm as shown in (b) and (c).
Then, shapes are correctly reconstructed as shown in (d).
In the next frame, two balls are reconstructed at further distance as shown in (e).

%Finally, a ball with small motion was reconstructed (Fig.~\ref{fig:caphand}). Since the velocity of the object is not 
%fast, we applied the phase based technique this time.
%As shown in (c), we can see that the phases are correctly estimated 
%Note that the round shape of the ball can be observed in (f).
%Also note that the $h(z)$ (Fig.~\ref{fig:caphand}(e)) is relatively constant compared to the {\it light flows} shown in (b,c),
%because variations of light flows in (b) and (c) are ``canceled out'' by the division.
%As shown in middle column, we can see that the phases are correctly estimated 
%even when the pattern was blurred by motion. 
%Using the phase, depths are directly reconstructed as shown in right column.

\jptext{
インパクトのための実験。

・扇風機回転の復元

・ボールの復元

人体系→手を動かすとか

%８．もしも余力があれば→我々の2p1cの等間隔バージョンと比較（パターンが
%等間隔であるという知識しか使ってないのと、オーバーラップ前提なので、今
%回の手法とかなり近い前提）→誤接続ありと無しとで復元して、誤接続ありの
%場合は我々の勝ち、としたいところ
}

\section{Conclusion} %\knote{川崎}}

In this paper, we have proposed techniques to reconstruct the shape of fast 
moving objects which are captured with motion 
    blur of a projected pattern using the ratio of {\it light flows} of two projections.
With the proposed technique, 
the distances can be directly calculated from local displacement information.
Encoding and decoding of global positional information from the pattern is not required,
%the ratio is calculated directly without explicitly estimating
%the flows on the projector image plane, 
which is usually a difficult task.
We have presented two types of setups, \ie, single and two projectors configurations, to efficiently estimate the 
    {\it light flow} and the depth of an object. Our experimental results 
    demonstrate 
    that depth is actually recovered by the ratio of the {\it light flows} on
    several moving objects,
    such as planar board, rotating fan and thrown ball. %, which have strong motion blur.
% , are successfully reconstructed by our technique.

\fnoteIII{
There are, of course, limitations in the proposed method, 
such as the spatial resolution becoming
as coarse as the line intervals. 
%Since we use line intervals with blur, 
%the depth resolution becomes, at best, as large as the line intervals. 
However, this is the first technique
which achieves depth estimation only from a flow of projected pattern,
which is observed as a blur with the best of our knowledge.
In addition, if we use a precision device like laser projector with
phase based analysis which can realize high precision pixel detection,
it can improve the accuracies and spatial resolution
in theory, 
which will be our future research.
%Our future research is to combine the technique to structured light or 
%photometric stereo techniques to increase accuracy.
}

\vspace{-0.2cm}
\section*{Acknowledgment}
\vspace{-0.2cm}
This work was supported in part by JSPS KAKENHI Grant No. 15H02779, 16H02849, 
MIC SCOPE 171507010 and MSR CORE12.
\vspace{-0.2cm}

%-------------------------------------------------------------------------
\clearpage
{\small
\bibliographystyle{ieee}
\bibliography{../bib/shortSTRING,../bib/JabRef,../bib/bibref,../bib/furukawa,../170721-cvpr-lightflow/egbib}
}

\end{document}